\documentclass[letterpaper, 10 pt, conference]{ieeeconf}  

\IEEEoverridecommandlockouts                              

\usepackage{graphicx} 
\usepackage{subfigure}
\usepackage{amsmath} 
\usepackage{amssymb}  
\usepackage{color}
\usepackage{colortbl}
\usepackage[dvipsnames]{xcolor}
\usepackage[linesnumbered,ruled,vlined]{algorithm2e}
\usepackage{threeparttable}
\usepackage{booktabs}

\title{\LARGE \bf
DOGE: An Extrinsic Orientation and Gyroscope Bias Estimation for Visual-Inertial Odometry Initialization
}

\author{Zewen Xu$^{1,2}$, Yijia He$^{3}$, Hao Wei$^{1*}$,  and Yihong Wu$^{1,2*}$
\thanks{*This work was supported by the National Natural Science Foundation of China under Grand No. 62202468 and the Youth Program of State Key Laboratory of Multimodal Artificial Intelligence Systems under Grand No. MAIS2024215. (Corresponding authors: Hao Wei and Yihong Wu.)}
\thanks{$^{1,2}$Zewen Xu, Hao Wei, and Yihong Wu are with the State Key Laboratory of Multimodal Artificial Intelligence Systems, Institute of Automation, Chinese Academy of Sciences, Beijing 100190, China. Zewen Xu, and Yihong Wu are also with the School of Artificial Intelligence, University of Chinese Academy of Sciences, Beijing 100190, China.  {\tt\small e-mail: \{xuzewen2020; weihao2019;  yihong.wu\}@ia.ac.cn}}
\thanks{$^{3}$ Yijia He is with TCL RayNeo, China. {\tt\small e-mail: heyijia2016@gmail.com}}
}

\begin{document}

\maketitle
\thispagestyle{empty}
\pagestyle{empty}

\begin{abstract}
Most existing visual-inertial odometry (VIO) initialization methods rely on accurate pre-calibrated extrinsic parameters. However, during long-term use, irreversible structural deformation caused by temperature changes, mechanical squeezing, etc. will cause changes in extrinsic parameters, especially in the rotational part. Existing initialization methods that simultaneously estimate extrinsic parameters suffer from poor robustness, low precision, and long initialization latency due to the need for sufficient translational motion. To address these problems, we propose a novel VIO initialization method, which jointly considers extrinsic orientation and gyroscope bias within the normal epipolar constraints, achieving higher precision and better robustness without delayed rotational calibration. First, a rotation-only constraint is designed for extrinsic orientation and gyroscope bias estimation, which tightly couples gyroscope measurements and visual observations and can be solved in pure-rotation cases. Second, we propose a weighting strategy together with a failure detection strategy to enhance the precision and robustness of the estimator. Finally, we leverage Maximum A Posteriori to refine the results before enough translation parallax comes. Extensive experiments have demonstrated that our method outperforms the state-of-the-art methods in both accuracy and robustness while maintaining competitive efficiency. 
\end{abstract}

\section{Introduction}
Visual-inertial odometry (VIO) is a technology that leverages measurements from both the inertial measurement unit (IMU) and camera to provide motion estimation and environment structure information for a wide range of downstream tasks. Benefiting from 
the compactness, low power consumption, and cost-effectiveness of these sensors, VIO has become a universal solution for augmented reality or virtual reality (AR/VR) applications, as well as navigation for micro aerial vehicles. 
In practical VIO applications, high precision and robustness are extremely important. For example, in AR applications, rapid high-quality initialization can ensure a natural and smooth virtual-real interaction experience, which is a key competitive factor for AR products \cite{concha2021instant}. In addition, since the extrinsic parameters between the camera and IMU are coupled into the constraint for solving VIO initialization, an accurate extrinsic parameter is meaningful for improving the initialization performance.

\begin{figure}
    \begin{minipage}{0.45\linewidth}
    \centering
    \includegraphics[width=0.9\linewidth]{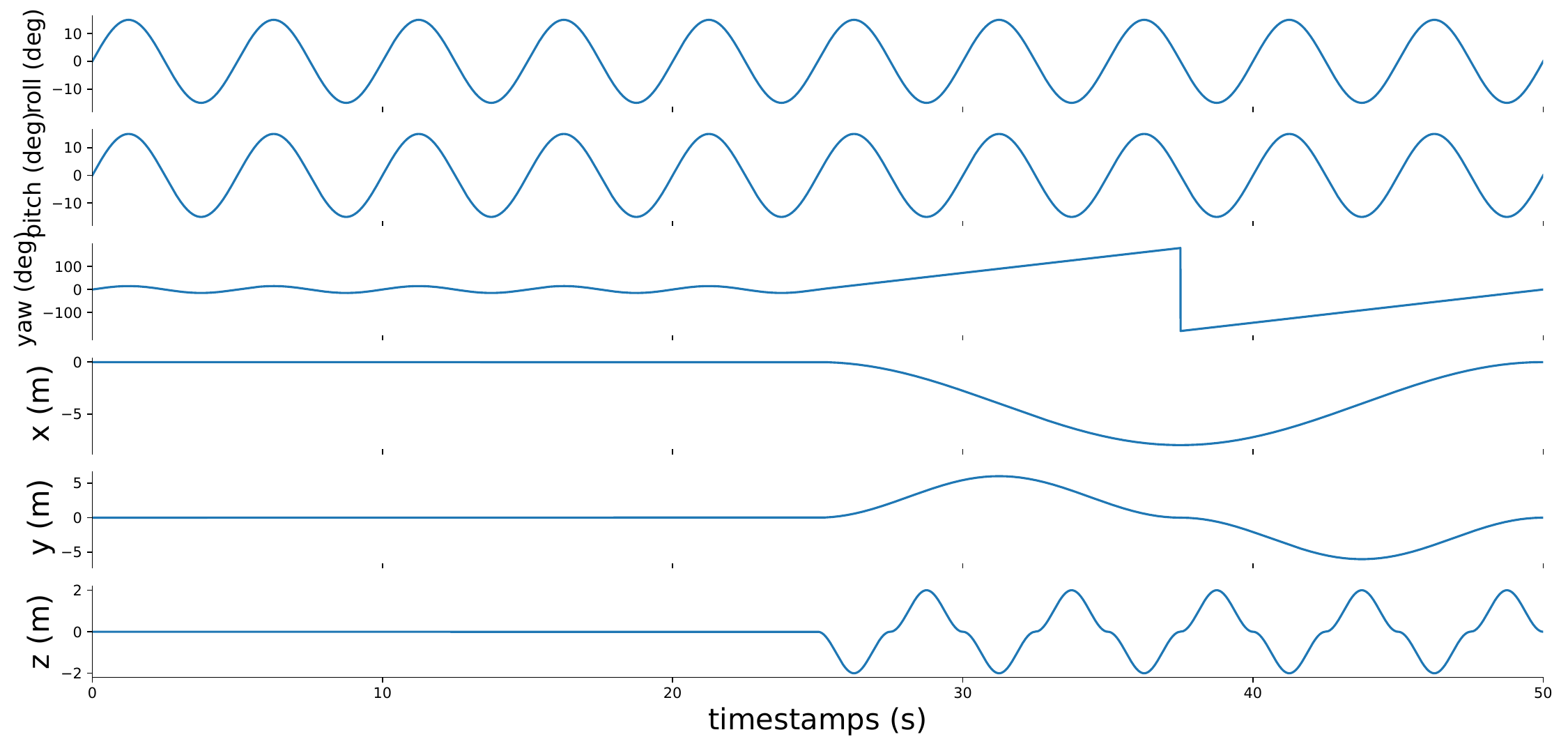}
    \end{minipage}
    \begin{minipage}{0.50\linewidth}
    \centering
    \includegraphics[width=0.8\linewidth]{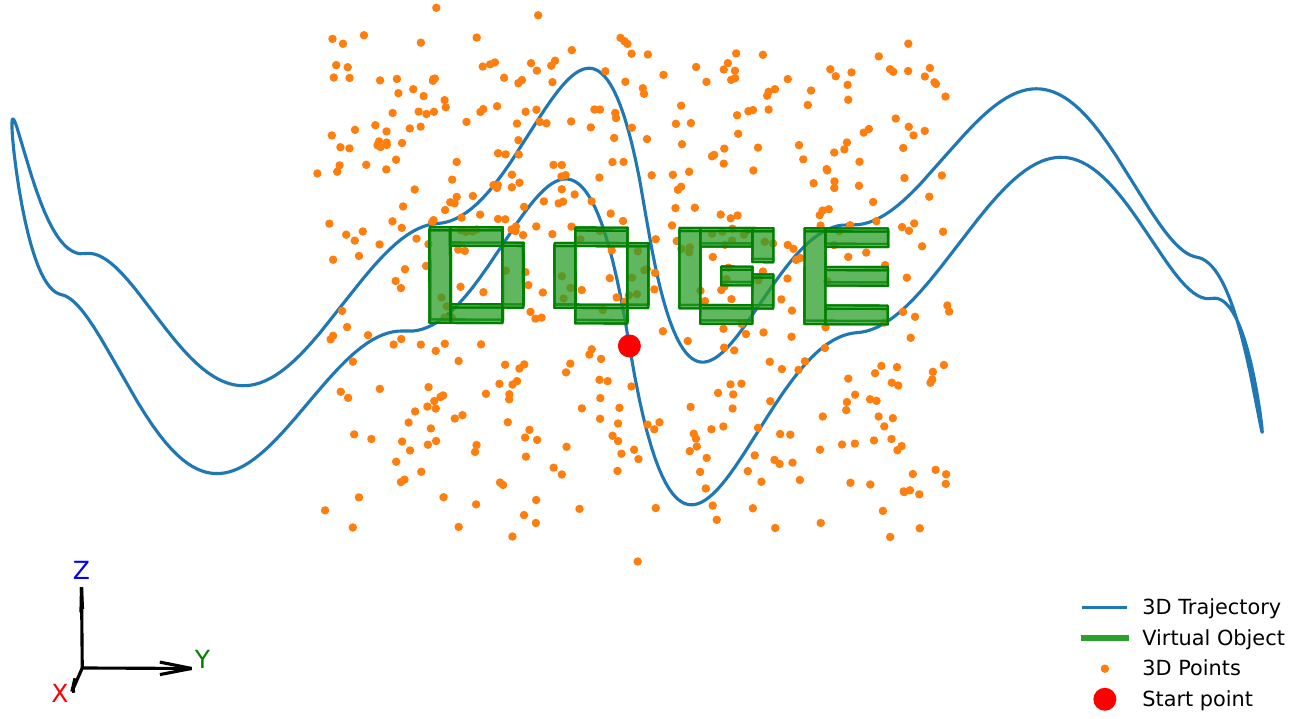}
    \end{minipage}
    \begin{minipage}{0.45\linewidth}
    \centering
    \includegraphics[width=0.99\linewidth]{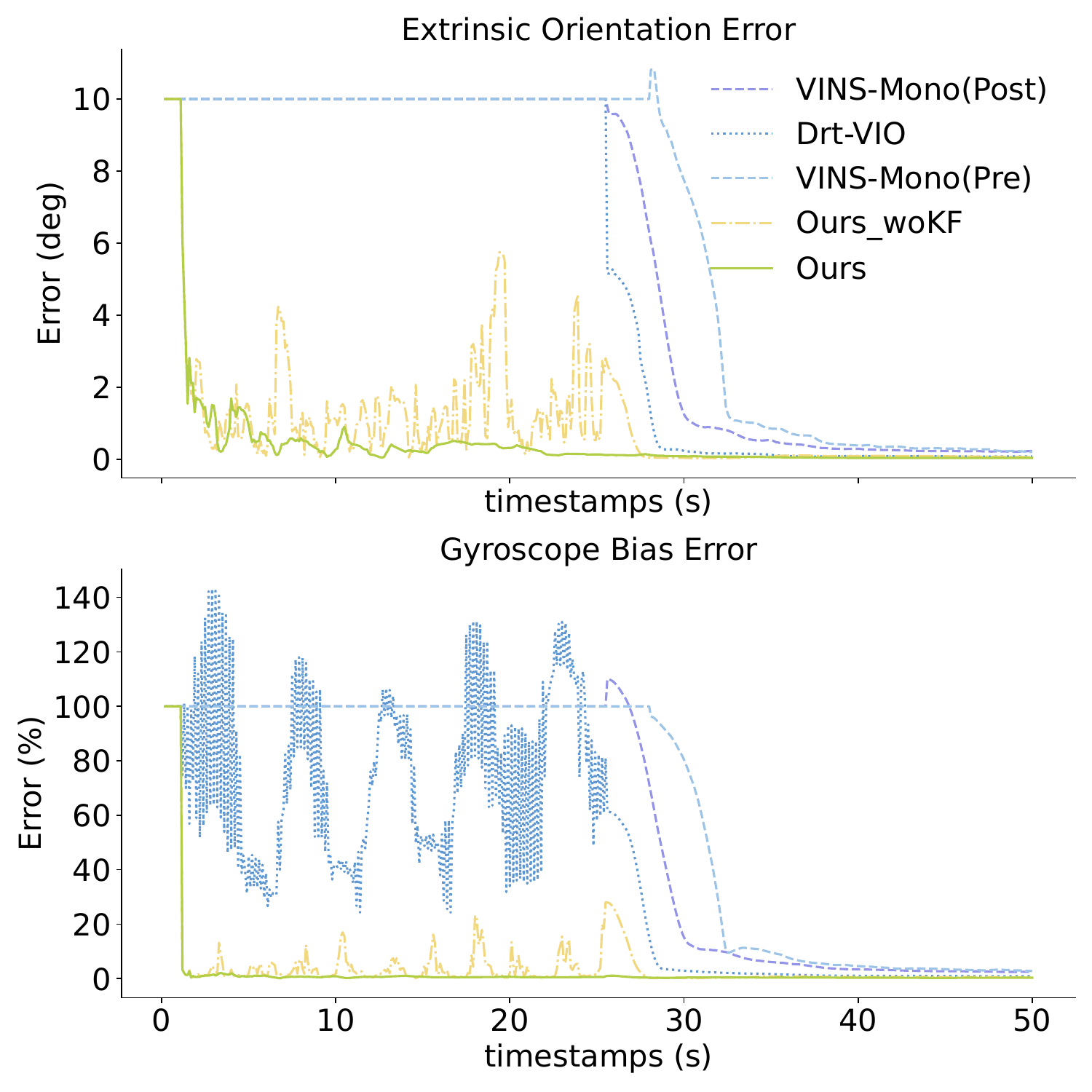}
    \end{minipage}
    \begin{minipage}{0.125\linewidth}
    \centering
        \includegraphics[width=1.0\linewidth]{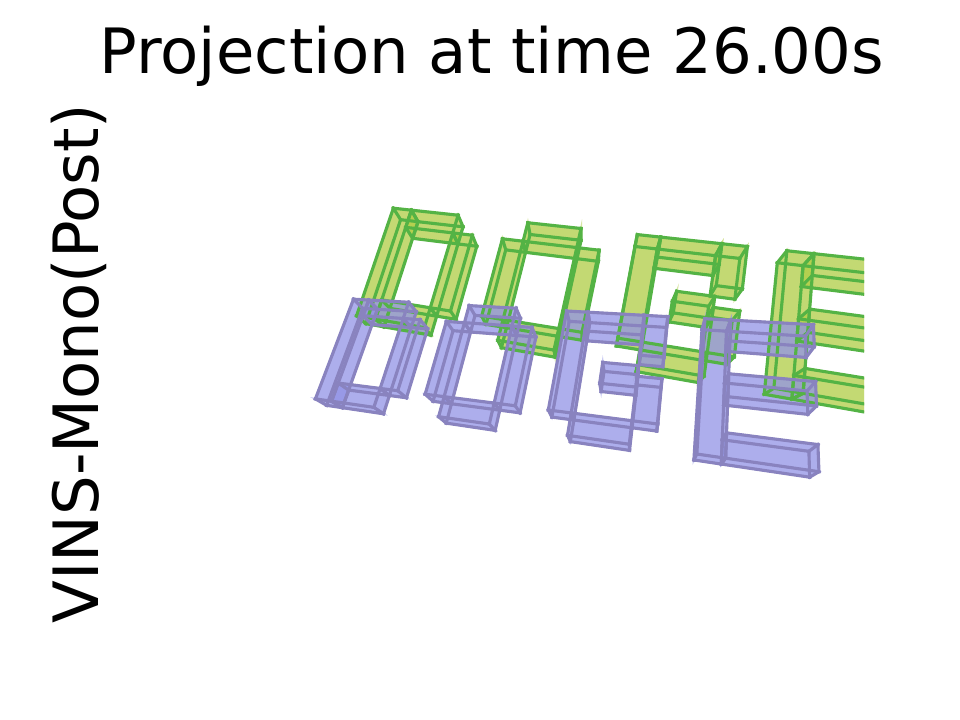}
        \includegraphics[width=1.0\linewidth]{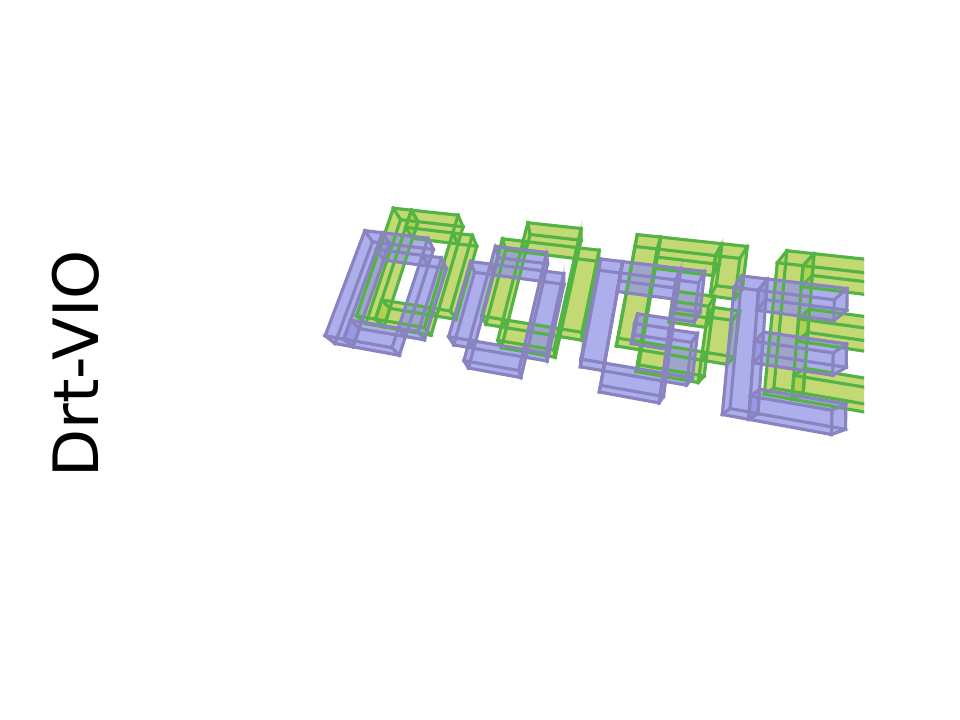}
        \includegraphics[width=1.0\linewidth]{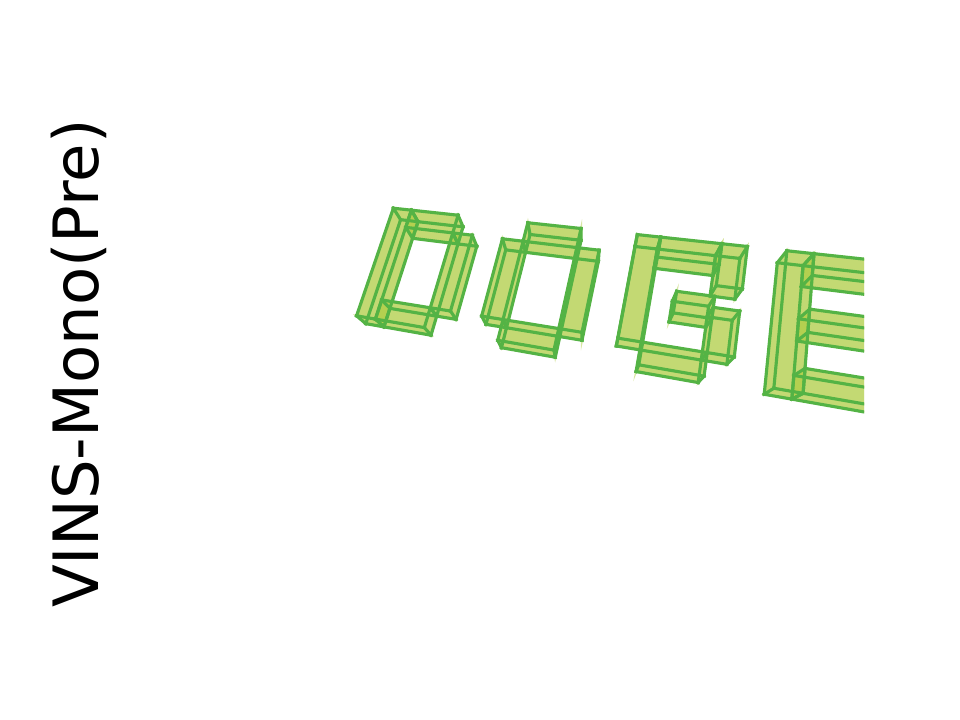}
        \includegraphics[width=1.0\linewidth]{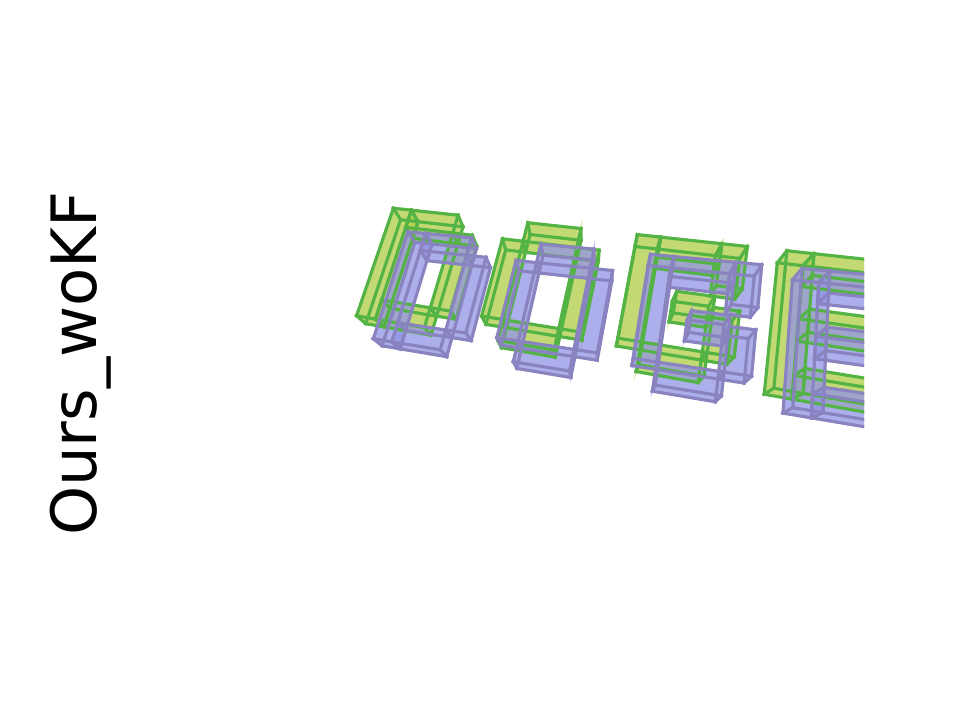}
        \includegraphics[width=1.0\linewidth]{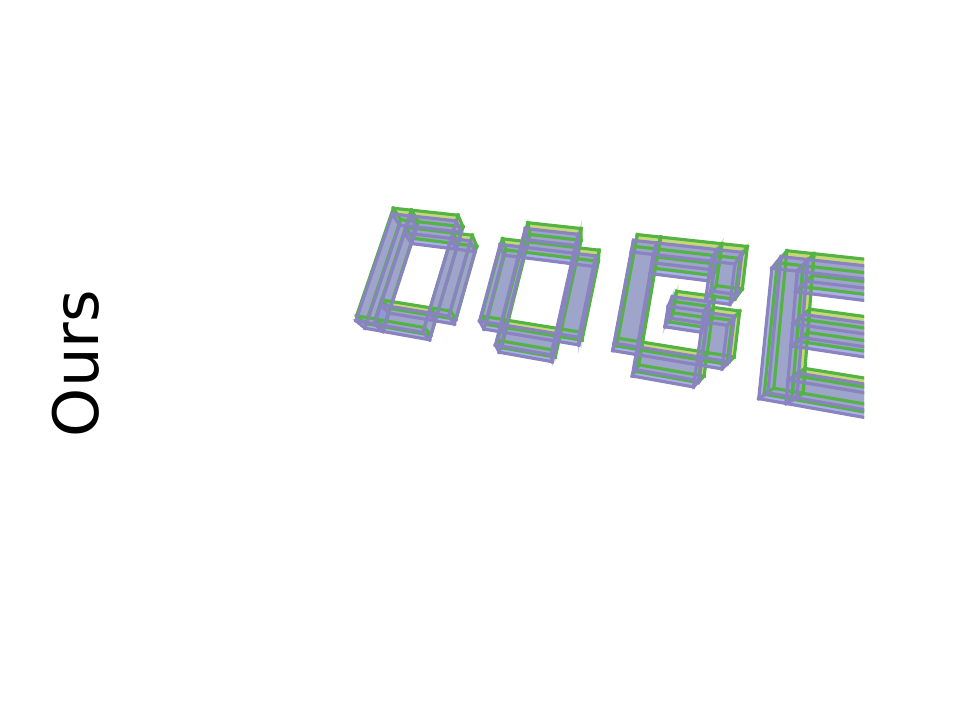}
    \end{minipage}
    \begin{minipage}{0.125\linewidth}
    \centering
        \includegraphics[width=1.0\linewidth]{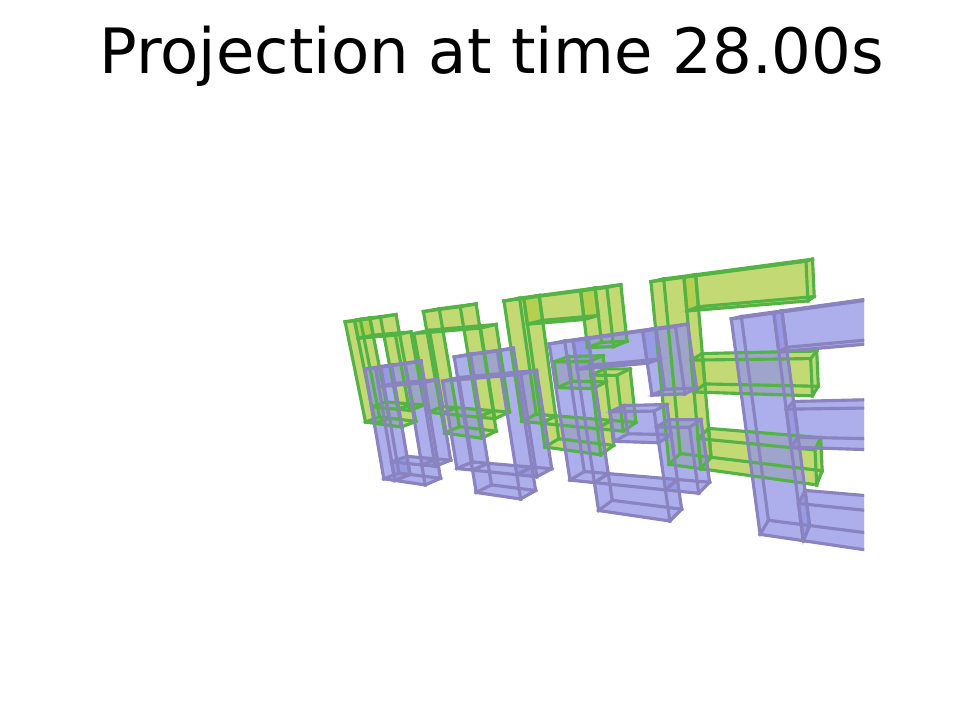}
        \includegraphics[width=1.0\linewidth]{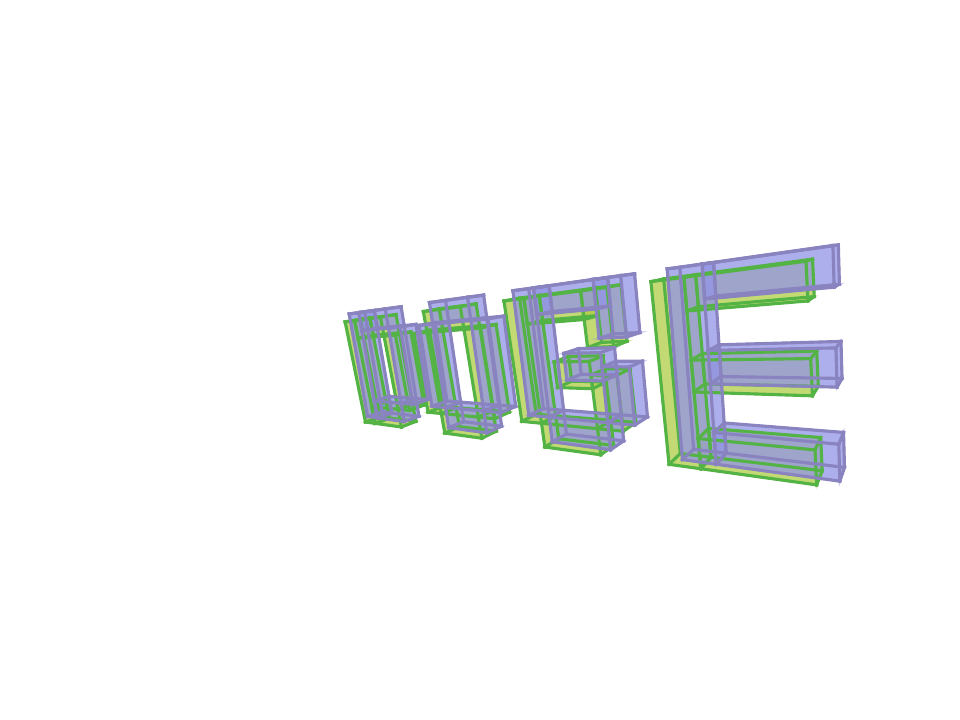}
        \includegraphics[width=1.0\linewidth]{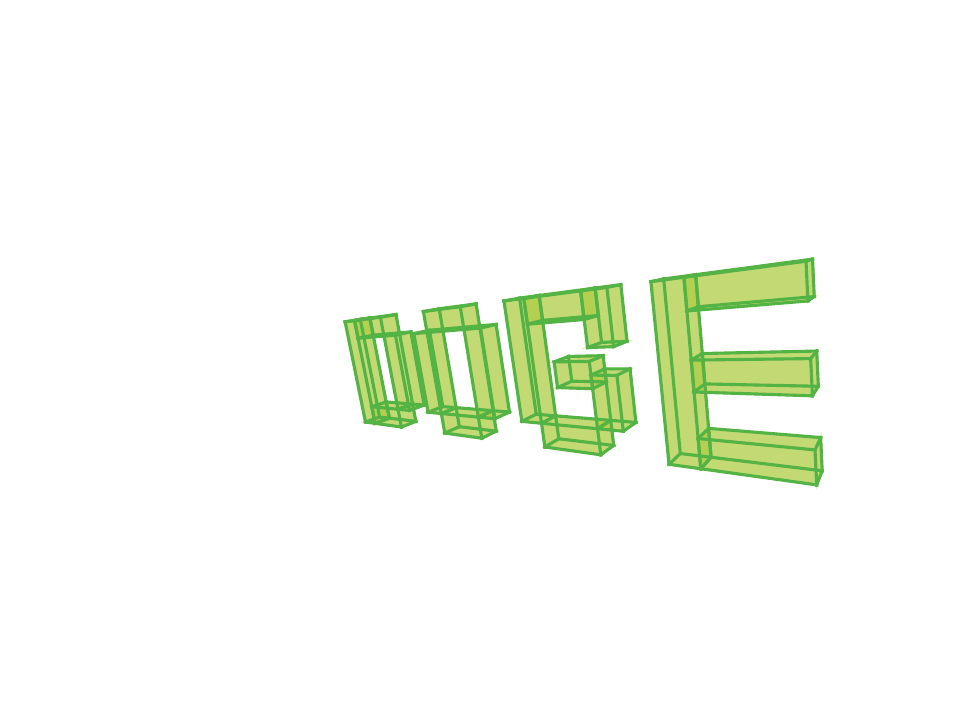}
        \includegraphics[width=1.0\linewidth]{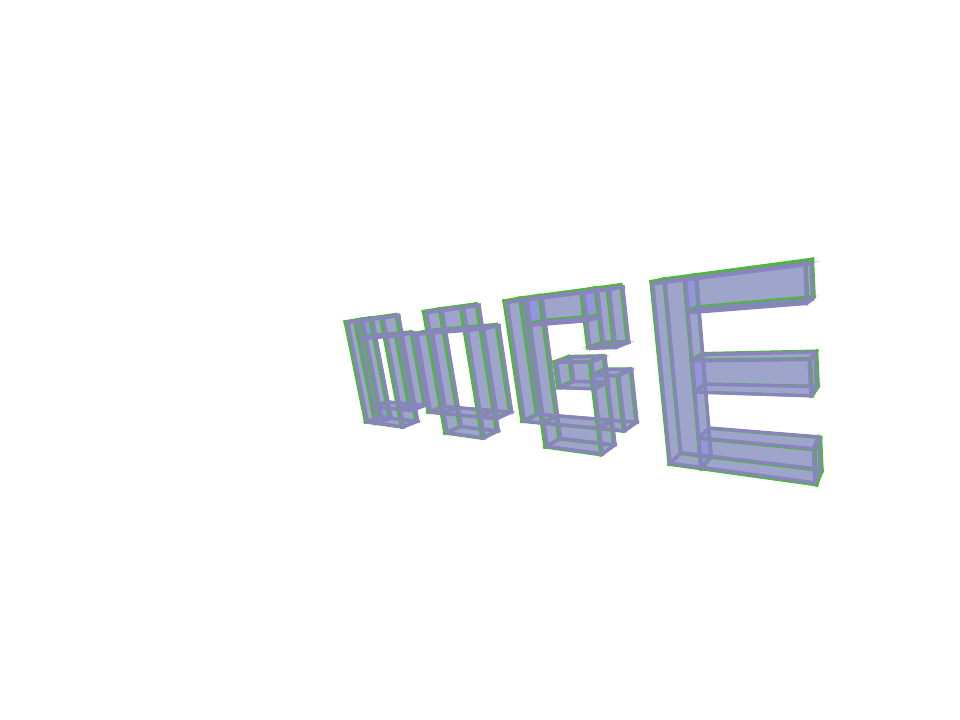}
            \includegraphics[width=1.0\linewidth]{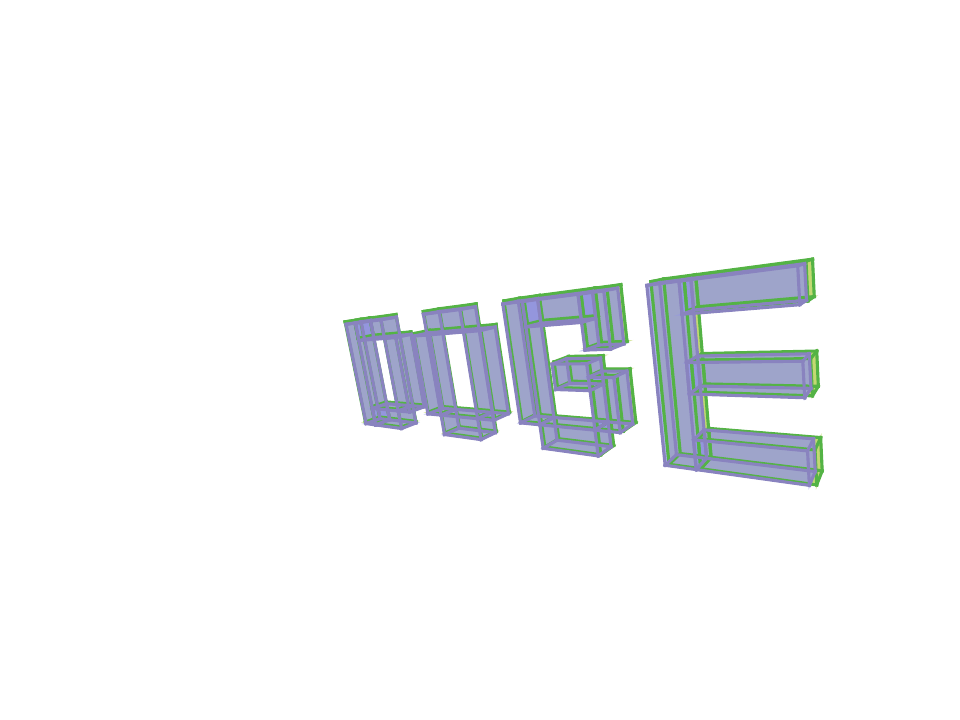}
    \end{minipage}
        \begin{minipage}{0.125\linewidth}
    \centering
        \includegraphics[width=1.0\linewidth]{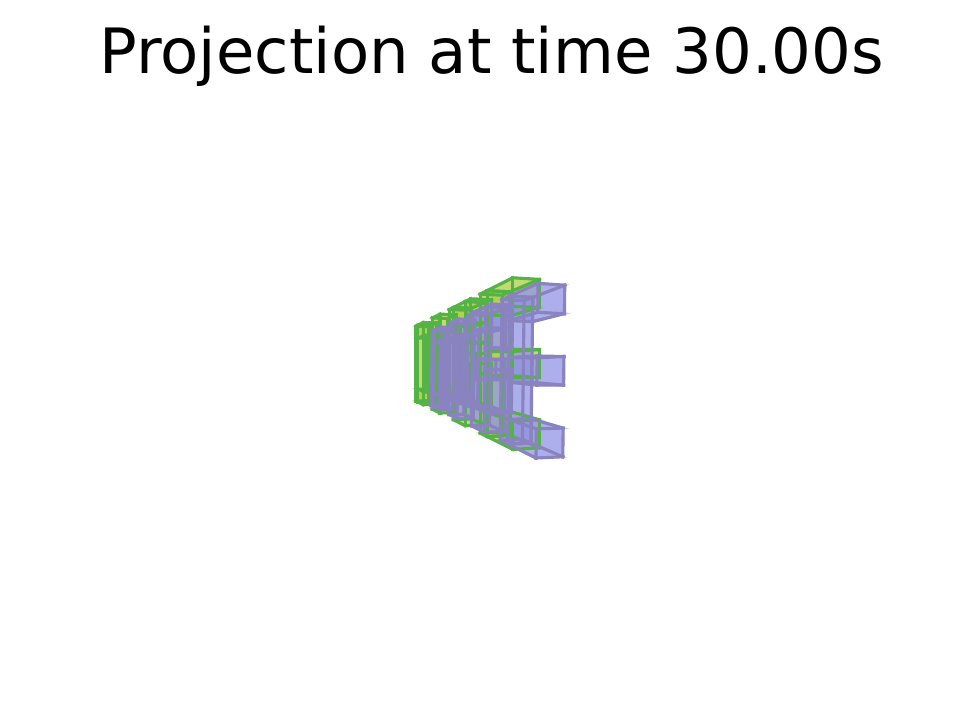}
        \includegraphics[width=1.0\linewidth]{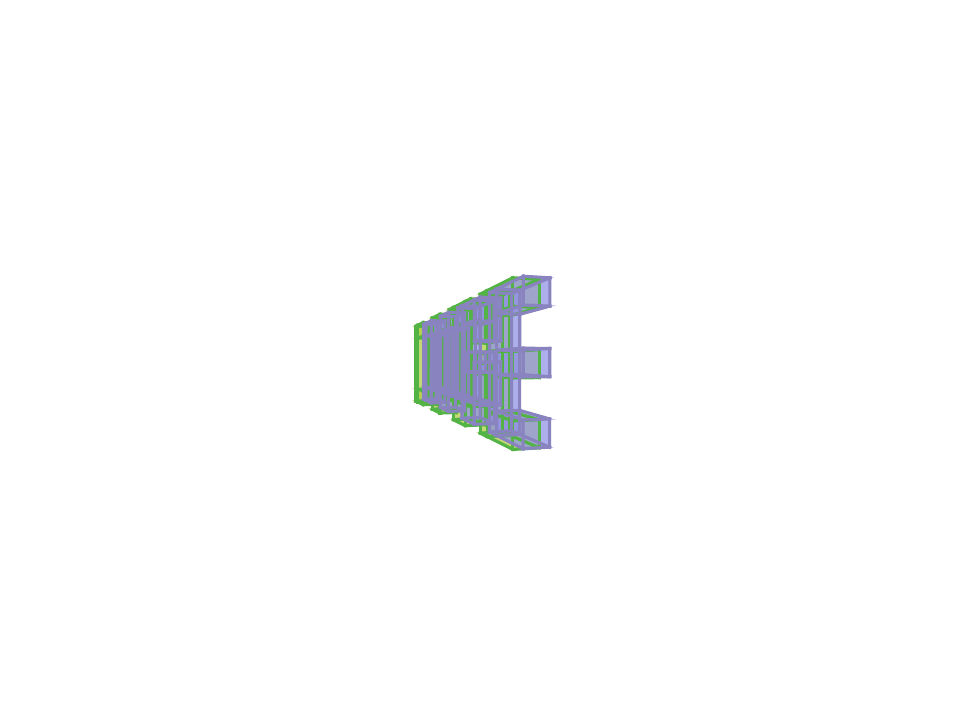}
        \includegraphics[width=1.0\linewidth]{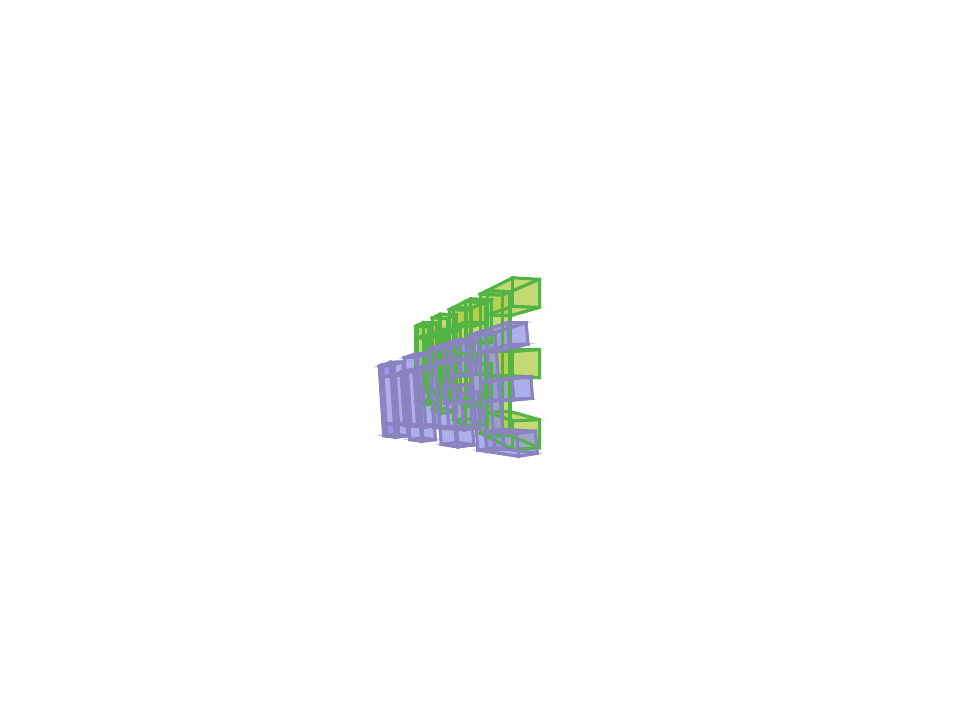}
        \includegraphics[width=1.0\linewidth]{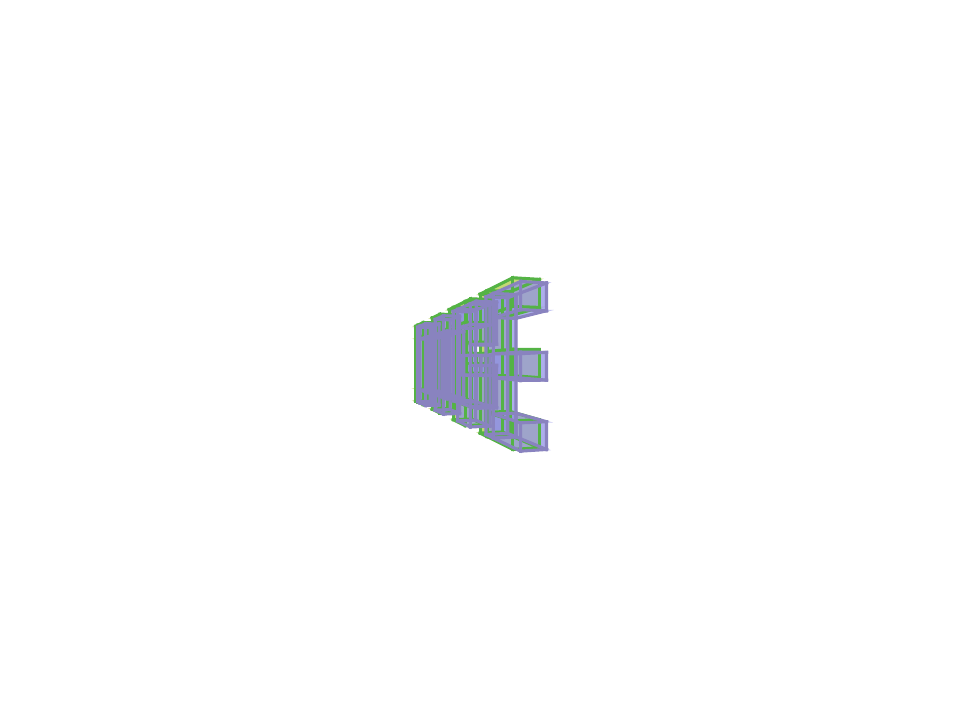}
            \includegraphics[width=1.0\linewidth]{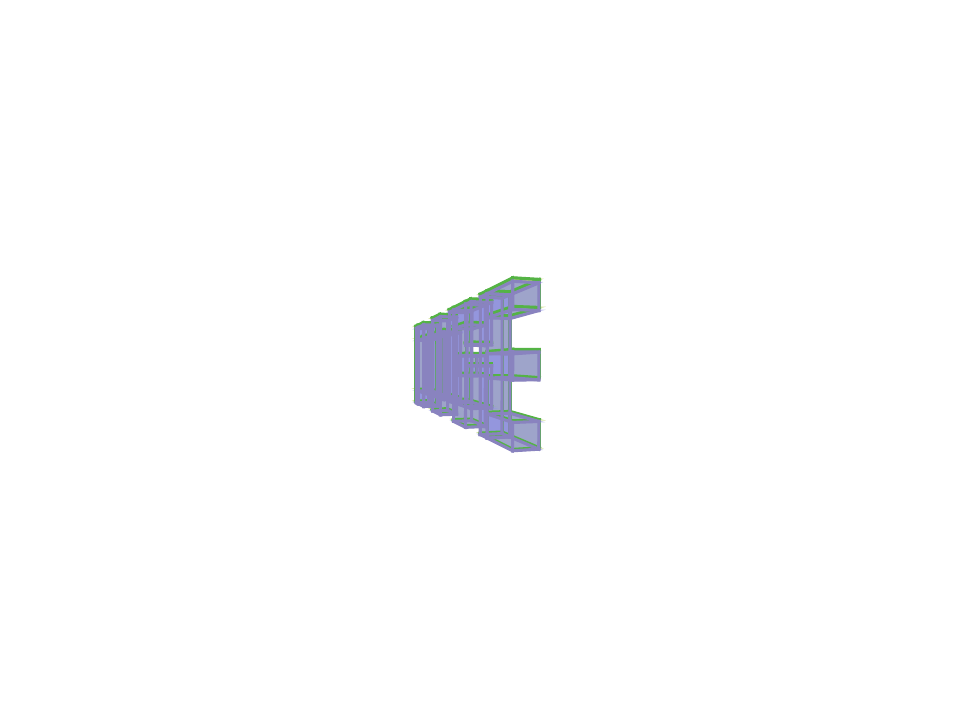}
    \end{minipage}
    \begin{minipage}{0.125\linewidth}
    \centering
        \includegraphics[width=1.0\linewidth]{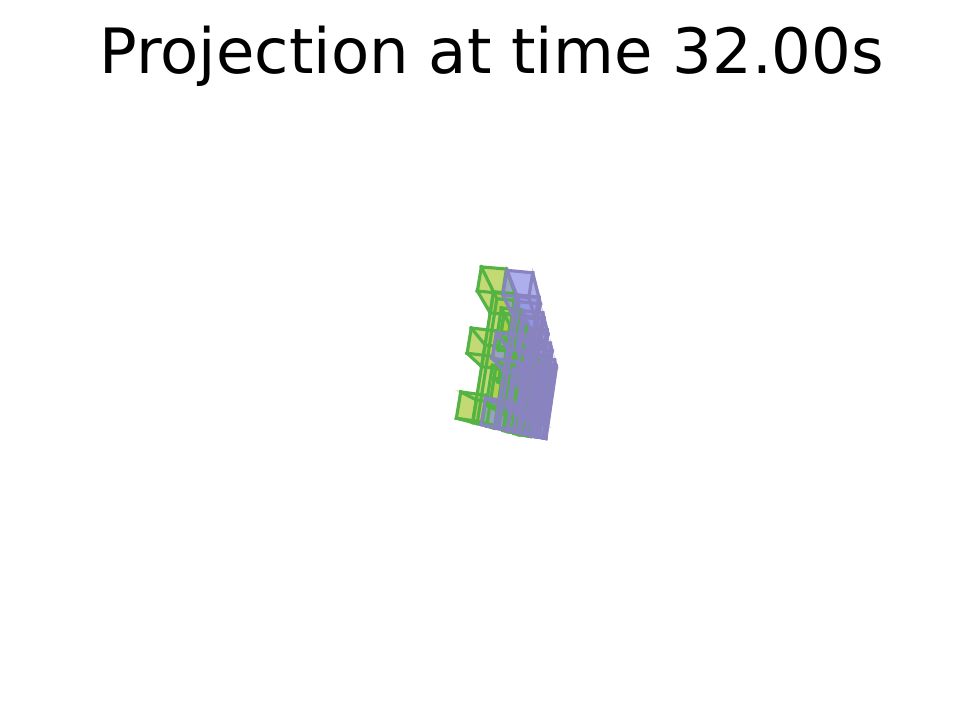}
        \includegraphics[width=1.0\linewidth]{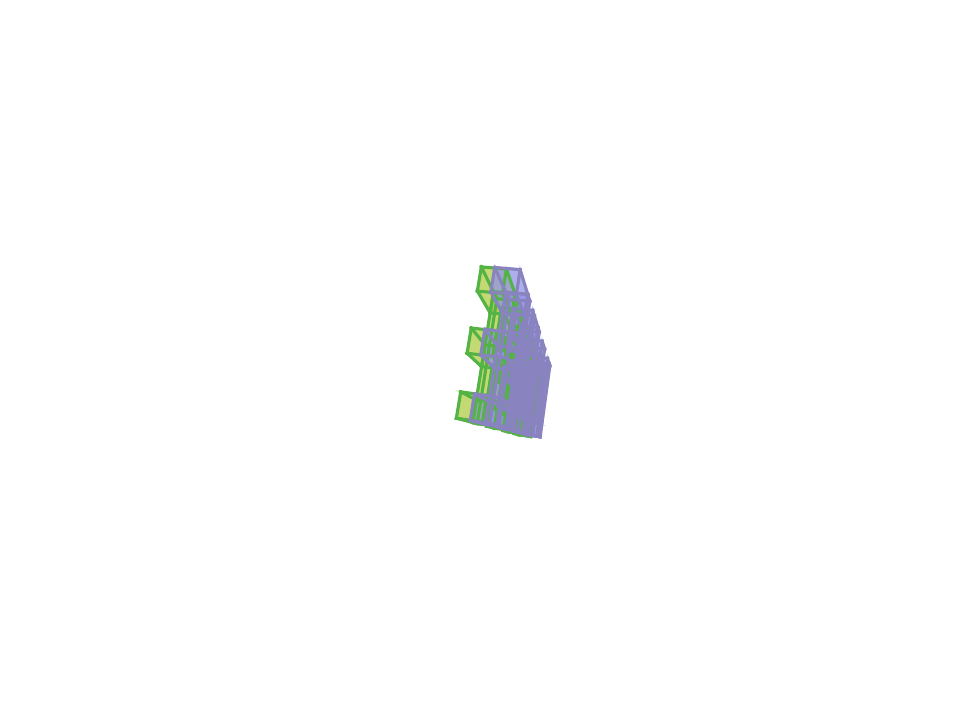}
        \includegraphics[width=1.0\linewidth]{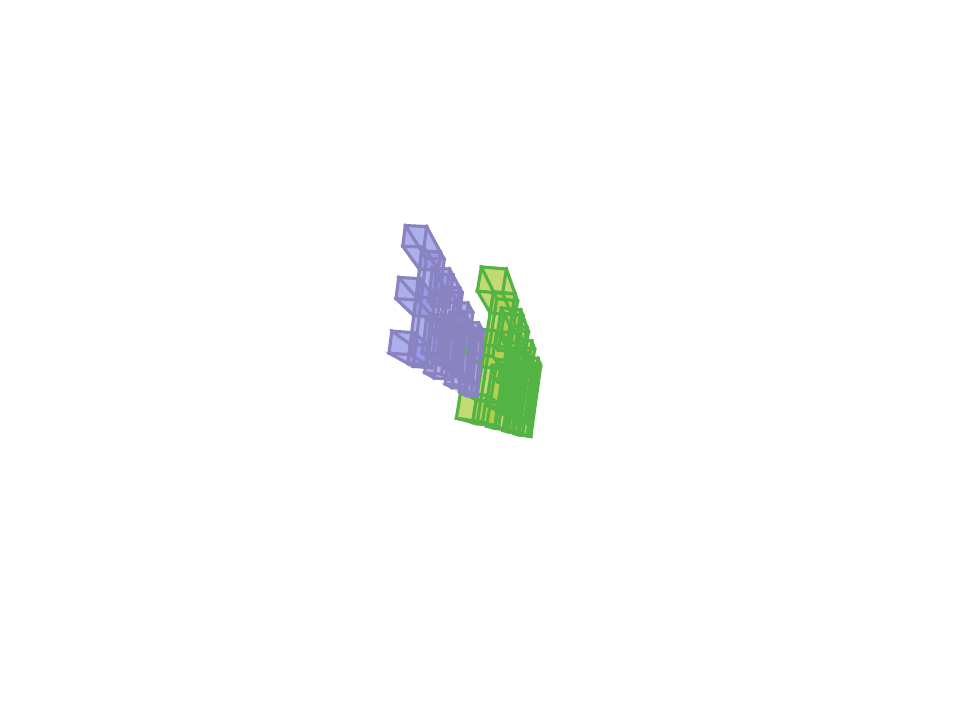}
        \includegraphics[width=1.0\linewidth]{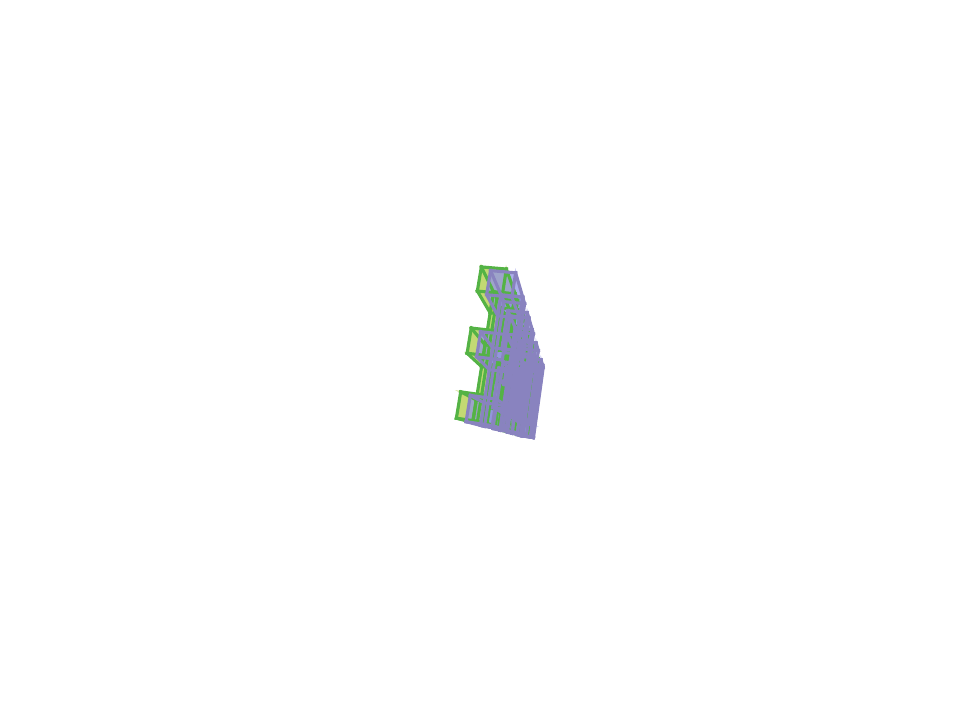}
        \includegraphics[width=1.0\linewidth]{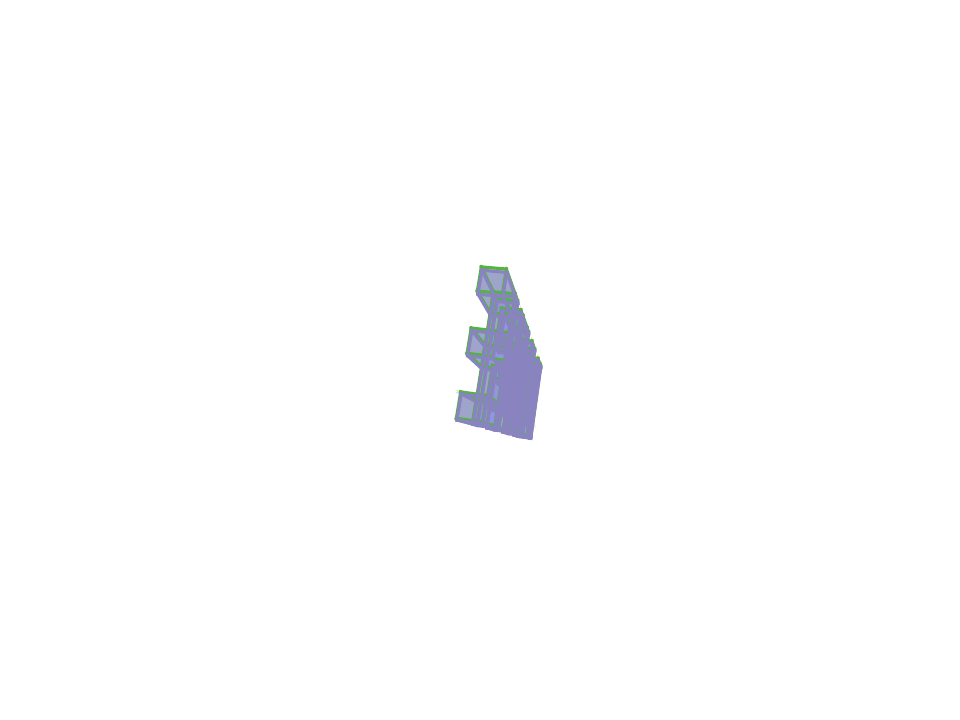}
    \end{minipage}
    \caption{\textbf{Illustration of the advantage of the proposed method for AR applications:} 
   The top row of images shows the simulation scene and the motion trajectory of the camera. In the first 25 seconds, the camera has only pure rotational motion. The detailed configuration can be found in Sec. \ref{sec:vio_test}. The \textbf{bottom left} image shows the estimation errors of rotational extrinsic parameters and gyroscope bias. 
   It can be seen that our method leverages the information from the pure rotation stage and gets a quicker convergence.
    The \textbf{bottom right} image shows the impact of initialization performance on VIO in AR applications. 
    The magnitude of the difference between the virtual object "DOGE" rendered with the estimated pose  (\textcolor{RoyalPurple}{purple}) and the ground truth (\textcolor{ForestGreen}{green}) in the figure indicates the performance of each method, clearly demonstrating the effectiveness of our method in terms of robustness, accuracy, and low latency.
    }
    \label{fig:AR_SIM_TEST}
    \vspace{-0.3in}
\end{figure}

 Many VIO initialization algorithms \cite{campos2021orb,he2023rotation,wang2024stereo} require accurate pre-calibrated extrinsic parameters, commonly obtained using professional offline tools like Kalibr \cite{furgale2013unified}. However, the mechanical structure is easily affected by long-term vibrations and temperature changes. For example, vibrations on drones and unmanned vehicles can cause structural deformation. Another example is that when wearing AR glasses, structural deformation may occur depending on the size of the user's face. Therefore, in actual use, the extrinsic parameters, especially the rotational part, may have undergone a non-negligible change compared to the factory-calibrated parameters. 
Usually, the translation part of the extrinsic parameters caused by deformation is very small, on the millimeter level, and it has little impact on the initialization performance \cite{qin2018vins}. So, we mainly focus on how to simultaneously solve the rotational extrinsic parameters in the initialization method.
 
 Currently, online extrinsic orientation estimation in the initialization stage can be divided into post-estimating \cite{Geneva2020ICRA,qin2017robust} and pre-estimating methods \cite{dong2012estimator,huang2018online}. Post-estimating methods estimate the initial state under an inaccurate guess of extrinsic orientation, then jointly optimize the initial states and extrinsic parameters within a nonlinear method, known as visual-inertial bundle adjustment (VI-BA). However, the convergence of VI-BA is influenced by the initial values. Inaccurate orientation and the resulting inaccurate initial states increase the risk of VI-BA failing to converge. On the other hand, pre-estimating methods first estimate the extrinsic orientation with a linear constraint, then use this result to initialize the states, and finally refine all parameters together. Compared to post-estimating methods, theoretically, these approaches can provide more accurate and robust initialization results in cases of poor extrinsic orientation. 
However, few pre-estimating methods estimate the extrinsic orientation and gyroscope bias together, resulting in worse outcomes \cite{qin2018vins,dong2012estimator}.
 Besides, most existing methods \cite{qin2018vins,Geneva2020ICRA,qin2017robust,dong2012estimator,huang2018online} require translation parallax or map information, which results in delayed rotational calibration (see Fig \ref{fig:AR_SIM_TEST}).

 According to \cite{ kneip2012finding,kneip2013direct}, the frame-to-frame rotation can be estimated independently of the translation component using the normal epipolar constraint (NEC). 
Recently, initialization methods\cite{he2023rotation,wang2024stereo} based on the NEC constraint that decouple rotation and translation have made significant progress in robustness, accuracy, and efficiency. However, they do not consider the impact of inaccurate extrinsic rotation on initialization performance.

To solve the problem of degraded initialization performance caused by changes in rotational extrinsic parameters due to deformation, we propose a novel method for joint extrinsic \textbf{O}rientation and \textbf{G}yroscope bias \textbf{E}stimation during the VIO initialization process in a rotation-translation \textbf{D}ecoupled way, namely DOGE.
Our main contributions are
\begin{itemize}
\item A joint constraint about extrinsic orientation and gyroscope bias is established in the NEC formulation, leading to a rotation-only solution to optimize these rotation-related parameters without translation parallax. 
\item A weighting strategy is proposed to incorporate visual observation uncertainty and gyroscope measurement uncertainty into NEC to enhance the estimation accuracy and robustness. A practical strategy for judging whether initialization is successful or not is introduced. 
\item A maximum a posteriori (MAP) estimation is formulated to refine the extrinsic orientation and gyroscope bias before enough translation excitation comes.
 \item Extensive experiments demonstrate the advantages of the proposed method both in accuracy and robustness than the state-of-the-art method while maintaining competitive efficiency.

\end{itemize}
\section{Notations and Preliminaries}
The following notations are used throughout this paper. Let $I_i$ and $C_i$ represent the IMU and camera frame at time-index $i$, respectively. $\bold{R}_{C_iC_j}\in SO(3)$ and $\bold{t}_{C_iC_j}\in \mathbb{R}^3$ denote the rotation and translation that take 3D points from $C_j$ to $C_i$. For a vector $\bold{v}$, $\lfloor\cdot\rfloor$ is skew-symmetric operator, $\Vert\cdot\Vert$ is the L2-norm, and $\Vert\bold{v}\Vert_{\bold{P}}$ is defined as $\bold{v}^T\bold{P}^{-1}\bold{v}$. Two operators on the manifold $\mathcal{M}$,  
$\boxplus: \mathcal{M}\times\mathbb{R}^n\to\mathcal{M}$ and $\boxminus: \mathcal{M} \times\mathcal{M}\to\mathbb{R}^n$, are used in this paper. For the Lie group $SO(3)$, they are:
\begin{equation}
    \bold{R}\boxplus\boldsymbol{\theta}=\bold{R}\mathrm{Exp}(\boldsymbol{\theta});\quad \bold{R^{\prime}}\boxminus\bold{R}=\mathrm{Log}(\bold{R}^{-1}\bold{R}^{\prime})
\end{equation}
where function $\mathrm{Exp}(\cdot): \mathfrak{so}(3)\to SO(3)$ is from Lie algebra to Lie group and function $\mathrm{Log}(\cdot): SO(3) \to \mathfrak{so}(3)$ is from Lie group to Lie algebra. We will revisit the rotation part of IMU preintegration and the main idea of NEC in the next subsections. We kindly refer the reader to \cite{forster2015imu} for the whole IMU model and to \cite{kneip2012finding,kneip2013direct} for deeper insight into the geometric interpretations of NEC.
\subsection{IMU Rotation Preintegration}
IMU measurements are affected by bias $\bold{b}$ and additive noise $\bold{n}$. Since the gyroscope bias is slowly time-varying, we assume keeping it constant as $\bold{b}^{I}_{g}$ in a sliding window during initialization. The raw gyroscope measurements represented in IMU frame, $\boldsymbol{\omega}^{I}_{m}$, are given by
\begin{equation}
    \boldsymbol{\omega}_{m,t}^{I}= \boldsymbol{\omega}^{I}_{t}+\bold{b}^{I}_{g}+\bold{n}^{I}_{g},
\end{equation}
where the additive noise is assumed as Gaussian white noise, $\bold{n}_{g}\sim\mathcal{N}(\bold{0}, \bold{\sigma}_g^2\bold{I} )$. Gyroscope bias is modeled as a random walk with Gaussian white noise derivatives, $\bold{n}_{\bold{b}_g}\sim\mathcal{N}(\bold{0}, \bold{\sigma}_{\bold{b}_g}^2 \bold{I})$. 
Given the gyroscope bias estimation $\bold{\widehat{b}}^{I}_g$, the orientation changes from time $i$ to time $j$ that can be obtained by integrating the gyroscope measurements
\begin{equation}
\boldsymbol{\widehat{\gamma}}^{I_{i}}_{I_{j}}=\prod\limits_{k=i}^{j-1}\mathrm{Exp}\left((\boldsymbol{\omega}^{I}_{m,k}-\bold{\widehat{b}}^{I}_{g})\Delta t\right),
\label{eq:preint}
\end{equation}
$\Delta t$ denotes the time interval between two adjacent IMU data. Eq. \eqref{eq:preint} is known as the standard rotation preintegration term. The effect of the error state $\delta\bold{b}^{I}_{g}$ on $\boldsymbol{\gamma}^{I_{i}}_{I_{j}}$ can be represented by a first-order approximation 
\begin{equation}
    \boldsymbol{\gamma}^{I_{i}}_{I_{j}} = \boldsymbol{\widehat{\gamma}}^{I_{i}}_{I_{j}}\mathrm{Exp}(\bold{J}^{\boldsymbol{\gamma}^{I_{i}}_{I_{j}}}_{\bold{b}^{I}_g}\delta\bold{b}^{I}_{g})
\end{equation}
where $\bold{J}^{\boldsymbol{\gamma}^{I_{i}}_{I_{j}}}_{\bold{b}^{I}_g}$ is the Jacobian of $\boldsymbol{\gamma}^{I_{i}}_{I_{j}}$ with respect to $\bold{b}^{I}_g$ can be calculated according to \cite{forster2015imu}.

\subsection{Normal Epipolar Constraints}
The core of the NEC constraint \cite{kneip2013direct} is to convert the standard epipolar constraint into the problem of coplanarity of the normal vectors of the epipolar planes so that rotation can be directly solved.
The epipolar plane about the $k$th point is spanned by its unit observation bearing vectors in two camera views $\boldsymbol{f}_i^k, \boldsymbol{f}_j^k$, and its normal vector is $ \bold{n}^k = \lfloor\boldsymbol{f}_i^k\rfloor_{\times}\bold{R}_{C_iC_j}\boldsymbol{f}_j^k$. All normal vectors of epipolar planes in two views are perpendicular to the translation vector $\bold{t}_{C_iC_j}$. For $n_{ij}$ co-visible 3D points, the energy function is constructed according to the coplanarity constraint:
\begin{equation}
\begin{aligned}
    E  &= \sum\limits_{k=1}^{n_{ij}}\Vert\bold{t}_{C_iC_j}^T {\bold{n}^{k}}\Vert^2 =\bold{t}_{C_iC_j}^T \sum\limits_{k=1}^{n_{ij}}({\bold{n}^{k}}{\bold{n}^{k}}^T) \bold{t}_{C_iC_j}\\
    &=\bold{t}_{C_iC_j}^T \bold{M}_{ij} \bold{t}_{C_iC_j}.
\end{aligned}
    \label{eq:point_cost_function}
\end{equation}
The above coplanarity problem can be solved by minimizing the minimum eigenvalue of $\bold{M}_{ij}$, $\lambda_{\bold{M}_{ij}}^{\mathrm{min}}$
\begin{equation}
    \bold{R}^{*}_{C_iC_j}=\underset{\bold{R}_{C_iC_j}}{\arg\min}\lambda_{\bold{M}_{ij}}^{\mathrm{min}}.
    \label{eq:NEC_constraints}
\end{equation}
Since $\bold{M}_{ij}$ is solely related to the rotation $\bold{R}_{C_iC_j}$, the rotation can be estimated in a decoupled way.

\section{The Proposed Initialization Framework}
Considering that rotation errors will affect the accumulation of translation vectors and ultimately affect trajectory accuracy, estimating gyroscope bias and the rotational part of extrinsic parameters is crucial in the initialization stage.
In this section, we first introduce our method for joint extrinsic orientation and gyroscope bias estimation. Then we accumulate rotation information within MAP estimation, leading to a precise and low-latency VIO initialization.

\subsection{Extrinsic Orientation and Gyroscope Bias Estimation}
\textbf{Objective function of DOGE}: First, we derive the NEC constraints about the extrinsic orientation and gyroscope bias. We represent the  gyroscope measurements in the camera frame,
\begin{equation}
    \boldsymbol{\omega}_{m,t}^{C}=\bold{R}_{CI}(\boldsymbol{\omega}^{I}_{t}+\bold{b}^{I}_{g}+\bold{n}^{I}_{g}),
\end{equation}
therefore, we can obtain 
 \begin{equation}
 \begin{aligned}
     \boldsymbol{\widehat{\gamma}}^{C_{i}}_{C_{j}} &=\prod\limits_{k=i}^{j-1}\mathrm{Exp}\left(\bold{\widehat{R}}_{CI}(\boldsymbol{\omega}^{I}_{m,k}-\bold{\widehat{b}}^{I}_{g})\Delta t\right) \\
     & = \bold{\widehat{R}}_{CI} \boldsymbol{\widehat{\gamma}}^{I_{i}}_{I_{j}} \bold{\widehat{R}}_{CI}^T,
 \end{aligned}
\label{eq:preint_c}
\end{equation}
where $\boldsymbol{\widehat{\gamma}}^{C_{i}}_{C_{j}}$ represents the rotation estimation from frame $C_i$ to frame $C_j$, which is also equal to $\bold{\widehat{R}}_{C_iC_j}$.
the first-order effect of the error state $\delta\bold{b}^{I}_{g}$ and $\delta\boldsymbol{\theta}_{CI}$ on $\boldsymbol{\gamma}^{C_{i}}_{C_{j}}$ can be represented
\begin{equation}
    \boldsymbol{\gamma}^{C_{i}}_{C_{j}} = \boldsymbol{\widehat{\gamma}}^{C_{i}}_{C_{j}}\mathrm{Exp}\left(\bold{J}^{\boldsymbol{\gamma}^{C_{i}}_{C_{j}}}_{\bold{b}^{I}_g}\delta\bold{b}^{I}_{g} + \bold{J}^{\boldsymbol{\gamma}^{C_{i}}_{C_{j}}}_{\boldsymbol{\theta}_{CI}}\delta\boldsymbol{\theta}_{CI}\right),
    \label{eq:integration_C}
\end{equation}
where $\bold{J}^{\boldsymbol{\gamma}^{C_{i}}_{C_{j}}}_{\bold{b}^{I}_g}$ and $\bold{J}^{\boldsymbol{\gamma}^{C_{i}}_{C_{j}}}_{\boldsymbol{\theta}_{CI}}$ denote the Jacobian of $\boldsymbol{\gamma}^{C_{i}}_{C_{j}}$ with respect to $\bold{b}^{I}_g$ and $\boldsymbol{\theta}_{CI}$, respectively, and can be calculated as below
\begin{equation}
\begin{aligned}
\bold{J}^{\boldsymbol{\gamma}^{C_{i}}_{C_{j}}}_{_{\bold{b}^{I}_g}}&=\bold{\widehat{R}}_{CI} \bold{J}^{\boldsymbol{\gamma}^{I_{i}}_{I_{j}}}_{\bold{b}^{I}_g}\\
\bold{J}^{\boldsymbol{\gamma}^{C_{i}}_{C_{j}}}_{\boldsymbol{\theta}_{CI}}&=\bold{\widehat{R}}_{CI} {\boldsymbol{\widehat{\gamma}}^{I_{i}}_{I_{j}}}^{T}  \bold{\widehat{R}}_{CI}^{T} - \bold{I}. 
\end{aligned}
\label{eq:jaccobian_C}
\end{equation}
Eq. \eqref{eq:preint_c} and Eq. \eqref{eq:jaccobian_C} ensure seamless integration of our method into popular VIO systems.
Moreover, the Jacobian about $\boldsymbol{\theta}_{CI}$ implies that as the IMU rotates more rapidly, the influence of extrinsic orientation becomes more pronounced. 

Combining Eq. \eqref{eq:NEC_constraints} and Eq. \eqref{eq:integration_C}, $\bold{M}_{ij}$ can be derived from gyroscope information and extrinsic orientation
\begin{equation}
\begin{aligned}
    \bold{M}_{ij}&=\sum\limits_{k=1}^{n}\left(\lfloor\boldsymbol{f}_{i}^{k} \rfloor_{\times}\boldsymbol{\widehat{\gamma}}^{C_{i}}_{C_{j}}\mathrm{Exp}\left(\bold{J}^{\boldsymbol{\gamma}^{C_{i}}_{C_{j}}}_{\bold{b}^{I}_g}\delta\bold{b}^{I}_{g} + \bold{J}^{\boldsymbol{\gamma}^{C_{i}}_{C_{j}}}_{\boldsymbol{\theta}_{CI}}\delta\boldsymbol{\theta}_{CI}\right)\boldsymbol{f}_{j}^{k}\right)\\
    &\quad\left(\lfloor\boldsymbol{f}_{i}^{k} \rfloor_{\times}\boldsymbol{\widehat{\gamma}}^{C_{i}}_{C_{j}}\mathrm{Exp}\left(\bold{J}^{\boldsymbol{\gamma}^{C_{i}}_{C_{j}}}_{\bold{b}^{I}_g}\delta\bold{b}^{I}_{g} + \bold{J}^{\boldsymbol{\gamma}^{C_{i}}_{C_{j}}}_{\boldsymbol{\theta}_{CI}}\delta\boldsymbol{\theta}_{CI}\right)\boldsymbol{f}_{j}^{k}\right)^T,
    \label{eq:Our_M}
\end{aligned}
\end{equation}
where $\delta\bold{b}^{I}_{g}$ and $\delta\boldsymbol{\theta}_{CI}$ need to be estimated. Note that $\bold{M}_{ij}$ is a positive semidefinite matrix, therefore, its minimum eigenvalue is non-negative. Let $\mathcal{K}$ denote the set of keyframe pairs that have enough co-visible features. The problem is formulated as follows:
\begin{equation}
    \delta\bold{b}^{I*}_{g}, \delta\boldsymbol{\theta}^{*}_{CI}=\underset{\delta\bold{b}^{I}_{g}, \delta\boldsymbol{\theta}_{CI}}{\arg\min}\sum\limits_{(i,j)\in\mathcal{K}}\lambda_{\bold{M}_{ij}}^{\mathrm{min}}.
    \label{eq:our_problem}
\end{equation}
According to \cite{kneip2012finding}, NEC can provide three independent constraints in a two-view setup with more than five co-visible points. Therefore, for the gyroscope bias and extrinsic orientation estimation, the size of $\mathcal{K}$ is at least $2$. 

When the initial value of extrinsic orientation is inaccurate the first-order approximation, Eq. \eqref{eq:integration_C}, will introduce a large linearization error. Therefore, we cannot get a satisfactory estimation in one step of solving when considering both extrinsic orientation and gyroscope bias jointly. In addition, different normal vectors about points in Eq. \eqref{eq:Our_M} should have different weights according to their position and observation noise. Due to these factors, we adopt an iterative reweighting approach to solve the minimization problem Eq. \eqref{eq:our_problem}. Concretely, we assume the observation noise of each point projection observes independent zero-mean Gauss distribution with $0.5$-pixel standard deviation, and propagate it through the unprojection function using the unscented transform \cite{uhlmann1995dynamic} to get full-rank 3D covariance matrices $\bold{\Sigma}_{i}^{k}$ and $\bold{\Sigma}_{j}^{k}$ of the unit bearing vectors. We propose the following weighting strategies.

$\boldsymbol{\lambda}$ \textbf{weighting strategy}:
Equation \eqref{eq:our_problem} can be directly transformed into the form of summing squared errors and introducing covariance $\sigma^{2}_{\lambda_{ij}}$ for each error term.
\begin{equation}
       \underset{\delta\bold{b}^{I}_{g}, \delta\boldsymbol{\theta}_{CI}}{\min}\sum\limits_{(i,j)\in\mathcal{K}}\left\Vert e_{ij}\right\Vert^{2}_{\sigma^{2}_{\lambda_{ij}}}, \text{ with } e_{ij}=\sqrt{\lambda_{\bold{M}_{ij}}^{\mathrm{min}}}.
    \label{eq:our_wls}
\end{equation}
For the calculation of the covariance $\sigma^{2}_{\lambda_{ij}}$, it can be intuitively seen from Eq. \eqref{eq:preint_c} and Eq. \eqref{eq:Our_M} that it is related to the covariance of image features and the covariance of IMU pre-integration $\bold{\Sigma}_{\boldsymbol{\gamma}^{I_i}_{I_j}}$. According to covariance propagation theory, we can derive the following equation.
\begin{equation}
\begin{aligned}
    \sigma_{\lambda_{ij}}^{2} = &\sum\limits_{k=1}^{n_{ij}}\left(\frac{\partial e_{ij}}{\partial \boldsymbol{f}_{i}^{k}}\bold{\Sigma}_{i}^{k}{\frac{\partial e_{ij}}{\partial \boldsymbol{f}_{i}^{k}}}^T +\frac{\partial e_{ij}}{\partial \boldsymbol{f}_{j}^{k}}\bold{\Sigma}_{j}^{k}{\frac{\partial e_{ij}}{\partial \boldsymbol{f}_{j}^{k}}}^T\right)\\
    &+ \frac{\partial e_{ij}}{\partial \boldsymbol{\gamma}^{I_i}_{I_j}}\bold{\Sigma}_{\boldsymbol{\gamma}^{I_i}_{I_j}}^{k}{\frac{\partial e_{ij}}{\partial \boldsymbol{\gamma}^{I_i}_{I_j}}}^T.
    \label{eq:weights_lambda}
\end{aligned}
\end{equation}
Since the error term in Eq. \eqref{eq:our_wls} is directly related to the eigenvalue $\lambda$, we call this weighting method the $\boldsymbol{\lambda}$ \textbf{weighting strategy}. However, in this way, 
feature outliers can not be culled and inevitably affect the accuracy and robustness of estimation,
because the accumulation of all features on two key-frames only contributes one residual term, unlike reprojection error where each feature is calculated independently. 
Therefore, a more meticulous strategy that can adjust the covariance of each feature as iteration optimization proceeds needs to be proposed.

\textbf{Feature pairs (FP) weighting strategy:} 
For the target of more meticulous weighting, we need to revisit the geometric interpretations of NEC.
Specifically, the minimal eigenvalue of $\bold{M}_{ij}$ means the sum of the square of the distance from the normal vector to the co-plane. Eq. \eqref{eq:our_problem} is equivalently represented as

\begin{equation}
\begin{aligned}
     &\underset{\delta\bold{b}^{I}_{g}, \delta\boldsymbol{\theta}_{CI}}{\min}\sum\limits_{(i,j)\in\mathcal{K}}\sum\limits_{k=1}^{n_{ij}}\Vert e_{ij}^k\Vert^2, \\
     \text{with} &\quad e_{ij}^k = \bold{v}_{ij}^T(\lfloor\boldsymbol{f}_{i}^{k} \rfloor_{\times}\boldsymbol{\gamma}^{C_{i}}_{C_{j}}\boldsymbol{f}_{j}^{k})=\bold{v}_{ij}^T(\bold{n}_{ij}^{k})
\end{aligned}
\label{eq:our_equal_problem}
\end{equation}
where $\bold{v}_{ij}$ is the eigenvector corresponding to the minimum eigenvalue of $\bold{M}_{ij}$. Compared to Eq. \eqref{eq:our_wls}, Eq. \eqref{eq:our_equal_problem} has finer-grained error terms. If we ignore the noise from $\bold{v}_{ij}$ and $\boldsymbol{\gamma}^{C_{i}}_{C_{j}}$, the variance of one error term in Eq. \eqref{eq:our_equal_problem} will be only related to one feature pair and can be obtained by 
\begin{equation}
\begin{aligned}
{\sigma_{ij}^{k}}^2 =\frac{\partial e^{k}_{ij}}{\partial \boldsymbol{f}_{i}^{k}}\bold{\Sigma}_{i}^{k}{\frac{\partial e^{k}_{ij}}{\partial \boldsymbol{f}_{i}^{k}}}^T +\frac{\partial e^{k}_{ij}}{\partial \boldsymbol{f}_{j}^{k}}\bold{\Sigma}_{j}^{k}{\frac{\partial e^{k}_{ij}}{\partial \boldsymbol{f}_{j}^{k}}}^T, \text{with}\\
\frac{\partial e^{k}_{ij}}{\partial \boldsymbol{f}_{i}^{k}} = -\bold{v}_{ij}^T\lfloor \boldsymbol{\gamma}^{C_{i}}_{C_{j}}\boldsymbol{f}_{j}^{k} \rfloor_\times, \quad \frac{\partial e^{k}_{ij}}{\partial \boldsymbol{f}_{j}^{k}} = \bold{v}_{ij}^T\lfloor \boldsymbol{f}_{i}^{k} \rfloor_\times \boldsymbol{\gamma}^{C_{i}}_{C_{j}}.
\end{aligned}
\label{eq:weights}
\end{equation}
Under this weighting strategy, we can conduct a Chi-square test for each error term
to guarantee the validity of the uncertainty model and avoid error weighting of outliers without serious information loss. After the Chi-square test, we get an inlier set of feature pairs, $\mathcal{I}_{ij}$.
According to Eq. \eqref{eq:our_equal_problem}, weighing $e^{k}_{ij}$ using $w_{ij}^{k}=\frac{1}{\sigma_{ij}^{k}}$ is equal to weighing $\boldsymbol{f}_{j}^{k}$,
\begin{equation}
    w_{ij}^{k} e^{k}_{ij} = \bold{v}_{ij}^T(\lfloor\boldsymbol{f}_{i}^{k} \rfloor_{\times}\boldsymbol{\gamma}^{C_{i}}_{C_{j}}(w_{ij}^{k}\boldsymbol{f}_{j}^{k})), \quad (i,j)\in \mathcal{I}_{ij}.
\end{equation}

Then we combine \textbf{FP weighting} with $\boldsymbol{\lambda}$ \textbf{weighting} strategy. 
Setting ${\boldsymbol{f}_{j}^{k}}^{\prime} = w_{ij}^{k}\boldsymbol{f}_{j}^{k}$ and ${\bold{\Sigma}_{j}^{k}}^{\prime} = {w_{ij}^{k}}^2 \bold{\Sigma}_{j}^{k}$ and replacing $\boldsymbol{f}_{j}^{k}$ and $\bold{\Sigma}_{j}^{k}$ with them in Eq. \eqref{eq:our_wls}, we obtain a new minimization problem that is less affected by outliers and modeling inaccuracies while considering the uncertainty of visual observations and IMU measurements:
\begin{equation}
\underset{\delta\bold{b}^{I}_{g}, \delta\boldsymbol{\theta}_{CI}}{\min}\sum\limits_{(i,j)\in\mathcal{K}}\left\Vert\sqrt{\lambda_{\bold{M}^{\prime}_{ij}}^{\mathrm{min}}}\right\Vert^{2}_{\sigma^{\prime2}_{\lambda_{ij}}}
\label{eq:Our_weighted_problem}
\end{equation}
Finally, we can solve Eq. \eqref{eq:Our_weighted_problem} using an iteratively reweighted least square (IRLS) algorithm. Specifically, in the first loop of IRLS, we do not conduct the weighting strategies as the initial values are inaccurate. Considering the subsequent weighting processes are influenced by the quality of the estimation in the first loop, we use a Cauchy robust function to improve the performance of the first loop's estimation. The minimal problems in each loop are solved by the Levenberg-Marquardt (LM) algorithm. 

\textbf{Failure Detection:} 
For the sake of solving efficiency, we do not ensure the accuracy of the solution through the inefficient random initialization method \cite{muhle2022probabilistic, kneip2013direct}. Instead, we propose a method to judge whether the solution is effective. When it fails, we will wait for new measurement information and start a new round of optimization.
We determine the quality of the solution by evaluating the pass rate of the Chi-square test. If the pass rate of total bearing vector pairs in the final loop is less than $\epsilon$, set to $0.8$ in our system
, we consider the initialization of the gyroscope bias and extrinsic rotation to have failed. 
if the estimation is successful, we will compute the covariance of parameters, $\bold{\Sigma}_{\bold{\overline{x}}}$, through the Fisher information matrix for the following processing. 
\subsection{DOGE-VIO Initialization}
We build MAP estimating and leverage an IESKF framework to refine the extrinsic orientation and gyroscope bias until sufficient translation parallax is achieved. 

DOGE outputs a Gaussian distribution $\bold{\overline{P}}_0 = \bold{\Sigma}_{\bold{\overline{x}}}$ for the unknown parameters in the initial sliding window $\bold{\overline{x}}_0=[\overline{\bold{b}}_{g0}^{IT},\overline{\boldsymbol{\theta}}_{CI0}^T]^T$.
Considering the random walk of gyroscope bias, we can get a Gaussian prior distribution and initial estimation for the parameters in the next sliding window:
\begin{equation}
\begin{aligned}
    \bold{\widehat{x}}_k &= \bold{\overline{x}}_{k-1}\\
    \bold{\widehat{P}}_k &= \bold{\overline{P}}_{k-1} + \begin{bmatrix}
    \sum\limits_{i=0}^{N-1}\frac{\bold{\sigma}_{\bold{b}_g}^{2}\Delta t_{i,i+1}^k}{N}\bold{I} & \bold{0}\\
    \bold{0} & \bold{0}
    \end{bmatrix},
\end{aligned}
\end{equation}
where $N$ denotes the size of the sliding window, $k$ denotes the sequence number of the sliding window,
$\Delta t_{i,i+1}^{k}$ denotes the time interval between adjacent keyframes in the $k$th window, specially, $\Delta t_{0,1}^{k}$ represents the time interval between the keyframe that just slid out and the oldest one in the window. The added noise means the average increased uncertainty, as we assumed the bias kept constant in a sliding window. 
Therefore, we can construct a prior constraints for $\bold{x}_k$, equivalently represented by $\delta\bold{x}_k^{\kappa}$:
\begin{equation}
\begin{aligned}
    \bold{x}_k\boxminus \bold{\widehat{x}}_k &= (\bold{\widehat{x}}_{k}^{\kappa}\boxplus\delta\bold{x}_{k}^{\kappa})\boxminus\bold{\widehat{x}}_k=\bold{\widehat{x}}_{k}^{\kappa}\boxminus\bold{\widehat{x}}_k+\bold{J^{\kappa}}\delta\bold{x}_{k}^{\kappa}\\
    &\sim\mathcal{N}(\bold{0},\bold{\widehat{P}}_k),
    \label{eq:prior}
\end{aligned}
\end{equation}
where the superscript $\kappa$ means the state of this parameters in $\kappa$th iteration, $\bold{J}^{\kappa}$ is the Jocabian of $(\bold{\widehat{x}}_{k}^{\kappa}\boxplus\delta\bold{x}_{k}^{\kappa})\boxminus\bold{\widehat{x}}_k$ with respect to $\delta\bold{x}_{k}^{\kappa}$ evaluated at zero. For the first iteration, $\bold{\widehat{x}}_{k}^{\kappa} = \bold{\widehat{x}}_{k}$ and $\bold{J}^{\kappa}=\bold{0}$. Combining the prior distribution in Eq. \eqref{eq:prior} with observation constraints Eq. \eqref{eq:Our_weighted_problem} yields the posterior distribution of $\bold{x}_k$, then the MAP of $\delta\bold{x}_{k}^{\kappa}$ is built as
\begin{equation}
\underset{\delta\bold{x}_{k}^{\kappa}}{\min}\left(\Vert\bold{\widehat{x}}_{k}^{\kappa}\boxminus\bold{\widehat{x}}_k+\bold{J^{\kappa}}\delta\bold{x}_{k}^{\kappa}\Vert_{\bold{\widehat{P}}_k}^2 + \sum\limits_{(i,j)\in\mathcal{K}}\left\Vert\sqrt{\lambda_{\bold{M}^{\prime}_{ij}}^{\mathrm{min}}}\right\Vert^{2}_{\sigma^{\prime2}_{\lambda_{ij}}}\right)
\end{equation}
After linearizing the observation constraints as $\sqrt{\lambda_{\bold{M}^{\prime}_{ij}}^{\mathrm{min}}} = \bold{z}_{ij}^{\kappa}+\bold{H}_{ij}^{\kappa}\delta\bold{x}_{k}^{\kappa}$, where $\bold{z}_{ij}^{\kappa}=\sqrt{\lambda_{\bold{M}^{\prime}_{ij}}^{\mathrm{min}}}\Big|_{\delta\bold{x}_{k}^{\kappa}=\bold{0}}$ and $\bold{H}_{ij}^{\kappa}$ is the Jacobian of $\sqrt{\lambda_{\bold{M}^{\prime}_{ij}}^{\mathrm{min}}}$ with respect to $\delta\bold{x}_{k}^{\kappa}$ evaluated at zero, this problem can be efficiently solved using the IESKF \cite{he2021kalman}. After gyroscope bias and extrinsic orientation initialization, we can initialize the initial velocity and gravity vector as done in \cite{he2023rotation}. Finally, VI-BA is used to refine all of parameters and estimate extrinsic translation and accelerometer bias.
\section{Experiments}
In this section, we evaluate the performances of the DOGE and the proposed entire initialization process on the popular EuRoC \cite{burri2016euroc} dataset. 
First, experiments comparing only initialization parameters are used to prove the effectiveness of our method.
Then, ablation experiments demonstrate the effectiveness of the weighting modules. Finally,  we integrate all the initialization methods into the VINS-Mono system, showing the impact of the proposed initialization process on enhancing the overall performance of the VIO system.
\subsection{Initialization Experiments}
In this subsection, the performance of each initialization method is carefully evaluated.
The classical and the state-of-the-art VI-initialization methods are used for comparison including the loosely coupled VI-initialization method in \cite{he2023rotation} (denoted as Drt), Drt with VI-BA formed a post-estimating method (denoted as Drt+VI-BA), Drt with the proposed weighting strategies and its VI-BA version (denoted as Drt-IRLS and Drt-IRLS+VI-BA), post-estimating initialization \cite{qin2017robust} used in VINS-Mono \cite{qin2018vins} without VI-BA (denoted as VINS), and the post-estimating in VINS-Mono (denoted as VINS+VI-BA). We do not compare the pre-estimating version of VINS-Mono in this section, cause it is hard to get a stable estimation with only 20 keyframes.
To comprehensively evaluate accuracy and robustness, we sampled 2393 data segments from the EuRoC dataset with sufficient rotational excitation. 
To analyze the impact of the rotational error of extrinsic parameters caused by deformation on initialization performance, we let the initial value of extrinsic parameters deviate from the true value to 0, 1, 5, 10, and 20 degrees.
For a fair comparison, we selected keyframes at $4$Hz, like \cite{he2023rotation}, for all initialization methods in this experiment. Besides, all algorithms use the same image processing operations like that in \cite{qin2018vins}, using sparse KLT algorithm \cite{lucas1981iterative} to track existing features and new corners are detected by \cite{shi1994good}. Each image maintains 150 points and outliers are culled using RANSAC \cite{fischler1981random}.

\begin{figure}[t]
\centering
\subfigure[$\bold{b}^I_g$ errors (\%)]{
\includegraphics[height=0.62\linewidth]{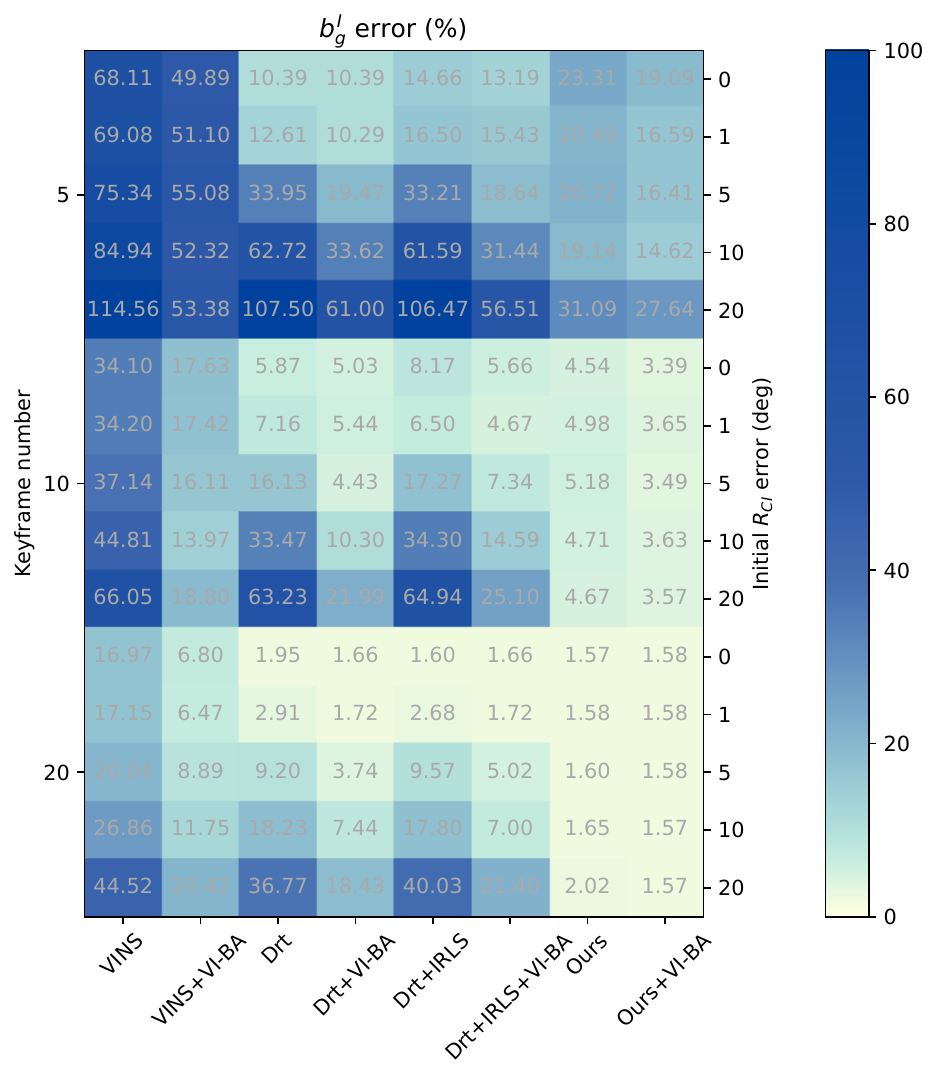}
}
\subfigure[$\bold{R}_{CI}$ errors (deg)]{
\includegraphics[height=0.62\linewidth]{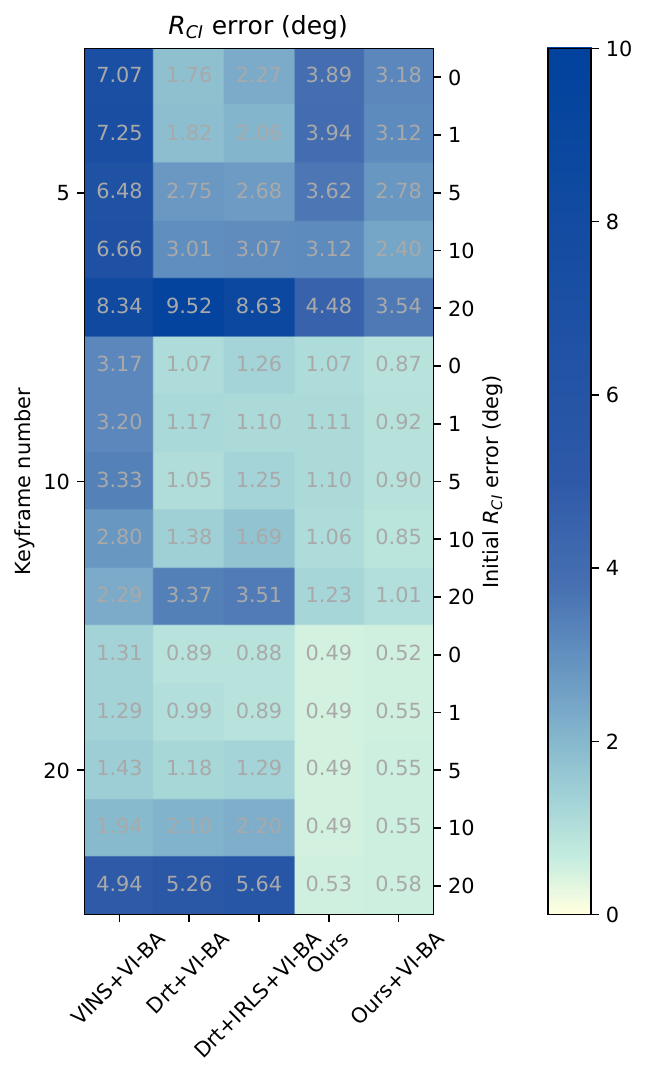}
}
\caption{\textbf{Accuracy evaluation:} The proposed method significantly outperforms the classical and state-of-the-art methods when the extrinsic orientation is poor.  }
\vspace{-0.2in}
\label{fig:acc_exp_bg}
\end{figure}
\begin{figure}[t]
\centering
\includegraphics[width=0.92\linewidth]{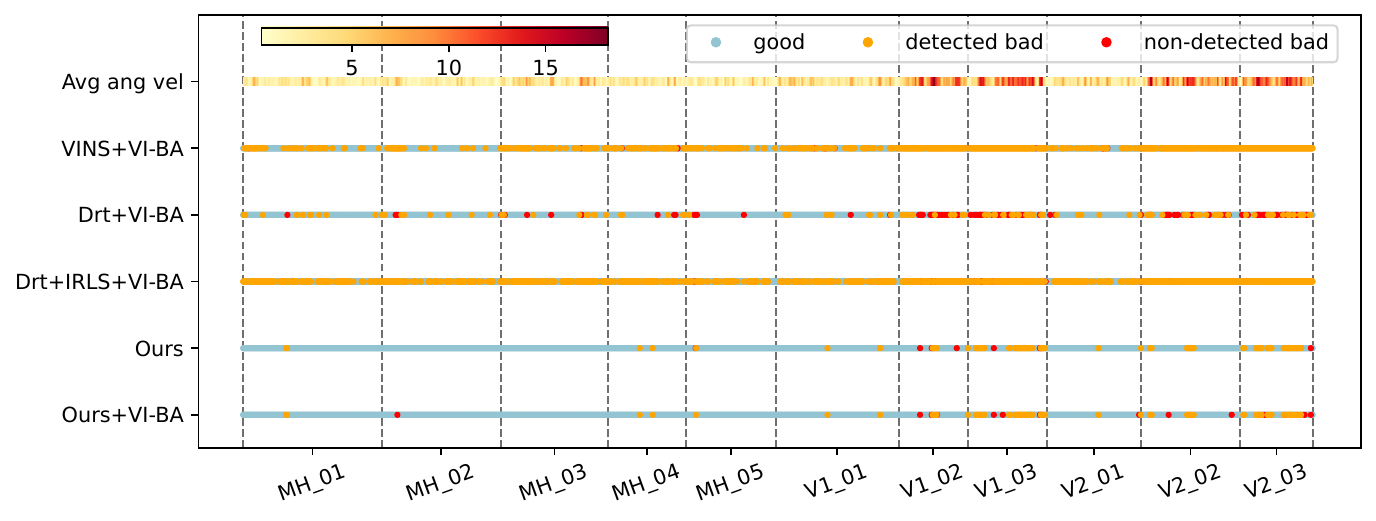}
\caption{\textbf{Robustness evaluation:} The top bar is colored according to the magnitude of the average angular velocity of the data segments (darker colors indicate higher magnitudes). The \textcolor{cyan}{cyan} points indicate data segments where a good estimation was obtained ($\bold{b}^I_g$ error is less than 50\% and the $\bold{R}_{CI}$ error is less than 5 degrees). The \textcolor{orange}{orange} points indicate data segments where a bad estimation is detected. The \textcolor{red}{red} points indicate segments where the method does not return a failure flag, but the $\bold{b}^I_g$ error exceeds 50\% or the $\bold{R}_{CI}$ error is larger than 5 degrees.}
\vspace{-0.1in}
\label{fig:robust_test}
\end{figure}
\begin{table}[t]
    \caption{Robustness Evaluation (\%)}
        \vspace{-0.1in}
    \centering
     \resizebox{0.95\linewidth}{!}{
    \begin{tabular}{lccccc}    
    \toprule
     & VINS+VI-BA & Drt+VI-BA & Drt+IRLS+VI-BA & Ours & Ours+VI-BA\\
    \midrule
       good &  55.87 & 88.05 & 34.77 & 94.40 & 94.07\\
       detected bad  & 43.38 &4.72 & 64.27 & 5.18& 5.18\\
       non-detected bad  & 0.75 & 7.23& 0.96& 0.42& 0.75 \\    
   \bottomrule
    \end{tabular}
    }
    \vspace{-0.1in}
    \label{tab:robust_test}
\end{table}
\begin{table}[t]
    \caption{Efficiency Evaluation (ms)}
        \vspace{-0.1in}
    \centering
       \begin{threeparttable}
     \resizebox{0.92\linewidth}{!}{
    \begin{tabular}{lccccc}    
    \toprule
    Module & VINS+VI-BA & Drt+VI-BA & Drt+IRLS+VI-BA & Ours & Ours+VI-BA\\
    \midrule
       SfM$^\dagger$   & 26.59 & - & - & - &- \\
       $\bold{b}_g$\&$\bold{R}_{bc}$ Est$^\star$ . & 0.53 & 3.32 & 5.00 & 13.18 & 13.18  \\
       Vel\&Grav Est.  & 0.21 & 0.81 & 0.81 & 0.81 & 0.81\\
       Point Tri. & 0.41 & 0.36 & 0.36 & 0.36 & 0.36\\
       VI-BA & 34.08 & 34.09 & 34.26 & -  & \textbf{29.51}\\
   \midrule
     Total Cost & 61.82 & 38.58 & 40.43 & \textbf{14.35} & 43.86 \\
   \bottomrule
    \end{tabular}
    }
    \begin{tablenotes}[flushleft]
     \footnotesize
        \item[$^\star$] Include consuming time for reintegration.
        \item[$^\dagger$] Structure from Motion
     \end{tablenotes}
    \end{threeparttable}
    \vspace{-0.2in}
    \label{tab:efficiency_test}
\end{table}

\textbf{Accuracy evaluation:} The accuracy of the rotational parameter estimation during initialization overall on the EuRoC dataset, under sliding window sizes of 5, 10, and 20 keyframes, is shown in Fig. \ref{fig:acc_exp_bg}. 
It can be seen that when the deformation angle is greater than 5 degrees, our method is significantly better than other methods in all configurations. And regardless of the magnitude of deformation or even no deformation, as long as the number of keyframes is appropriate, such as 10, our method achieves the optimal result. It is worth noting that the initialization parameters related to translation (such as scale, gravity direction, etc.) are not shown and compared here. On the one hand, it is because the quality of parameters related to rotation can represent the performance of all initialization parameters\cite{he2023rotation}. On the other hand, we will indirectly prove it through VIO experiments in Sec. \ref{sec:vio_test}.
\begin{figure}[t]
    \centering
    \includegraphics[width=0.89\linewidth]{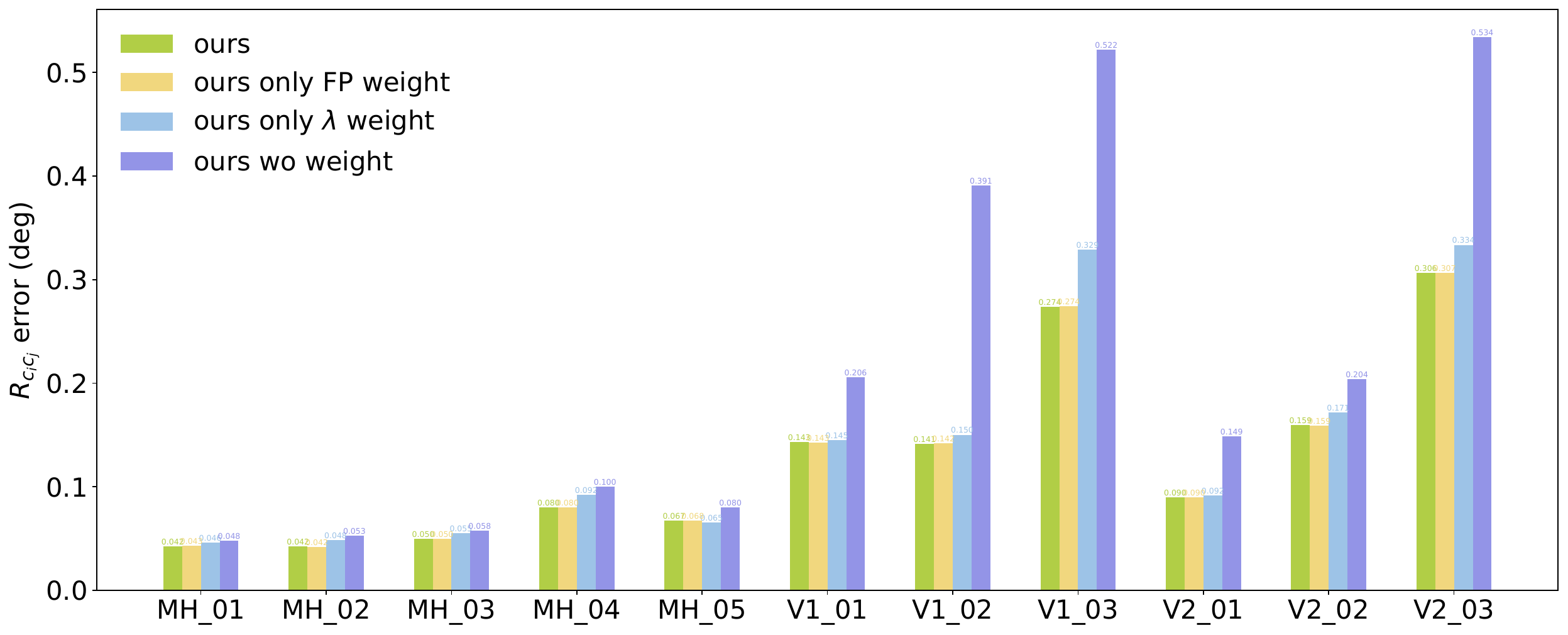}
     \vspace{-0.1in}
    \caption{\textbf{Ablation expriment}}
    \label{fig:ablation_test}
     \vspace{-0.1in}
\end{figure}
\begin{figure}[t]
    \centering
    \includegraphics[width=0.9\linewidth]{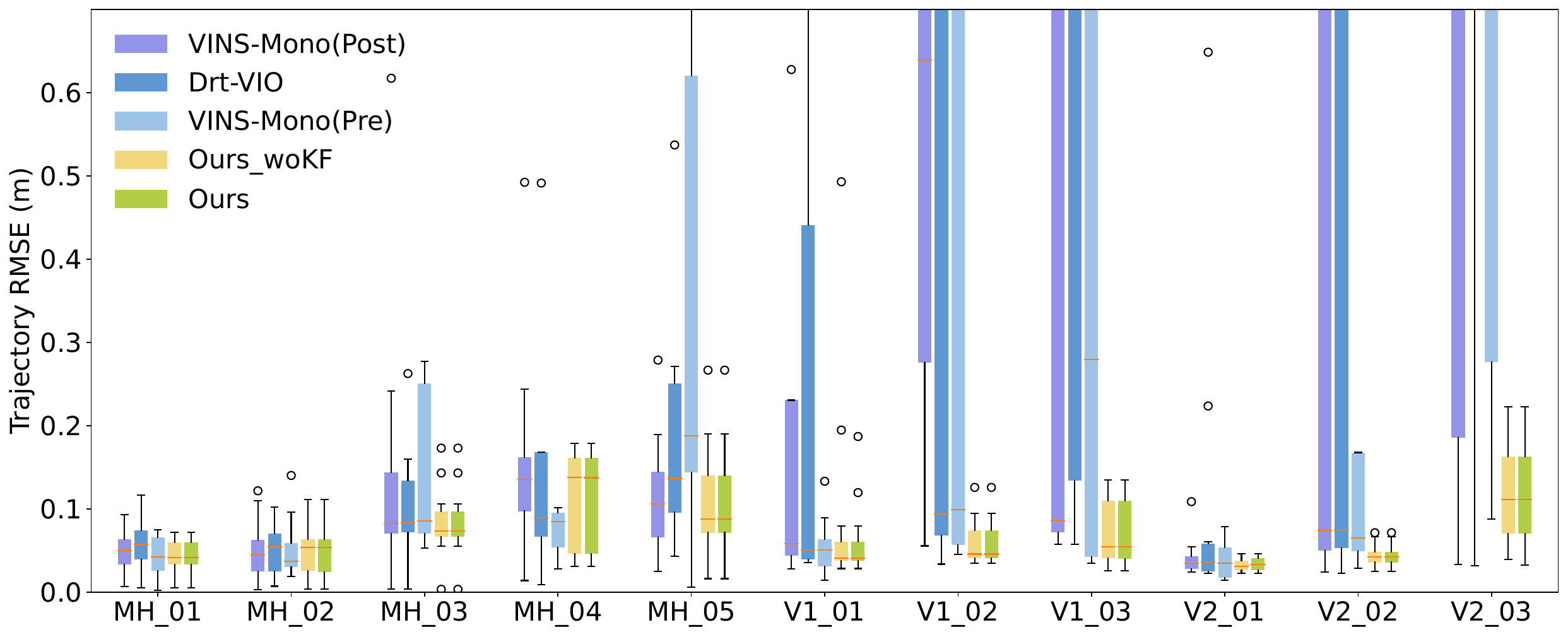}
     \vspace{-0.1in}
    \caption{\textbf{Test on EuRoC: } Each methods adopt the same configuration as VINS-Mono. Due to ignoring gyroscope bias, VINS-Mono(Pre) performs worse than VINS-Mono(Post). On contrary, the proposed method gets an outstanding performance on almost all sequences.}
    \label{fig:vio_euroc_test}
    \vspace{-0.1in}
\end{figure}
\begin{figure}[t]
    \centering
    \includegraphics[width=0.9\linewidth]{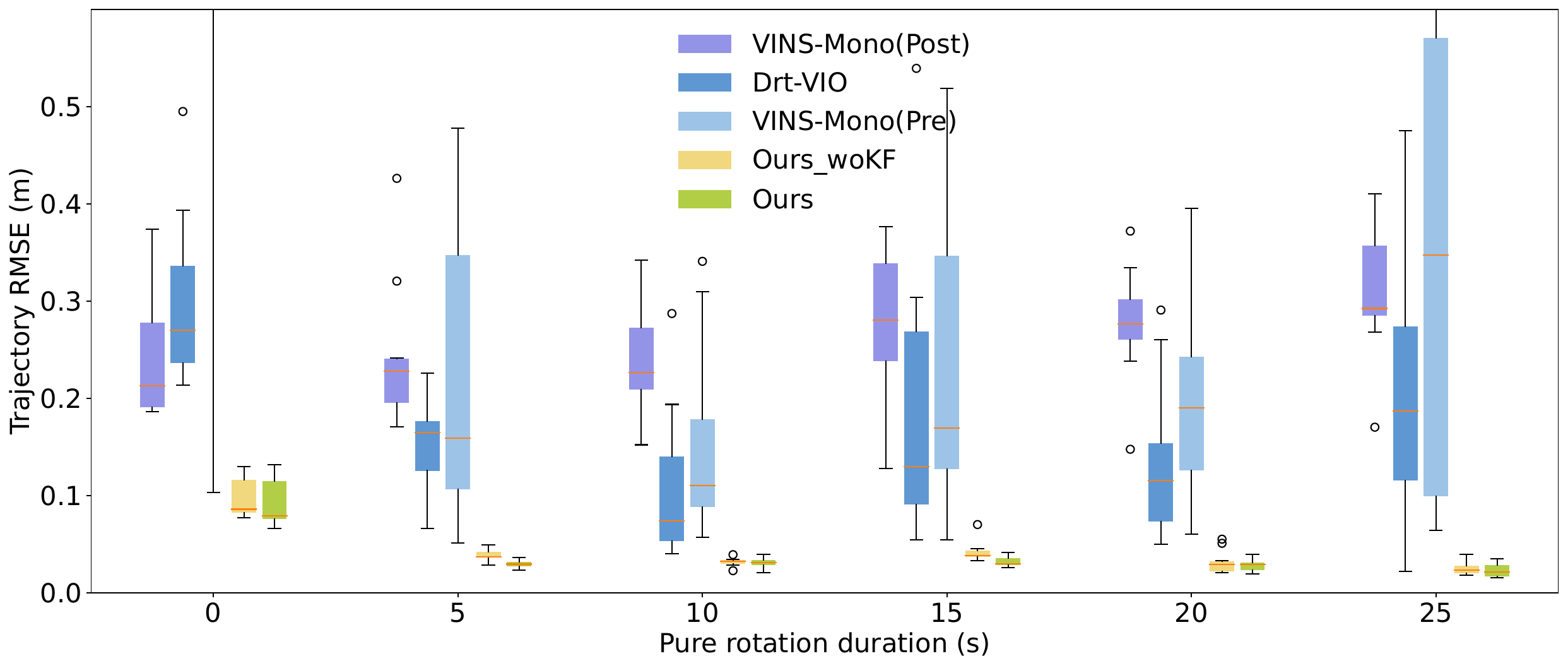}
     \vspace{-0.1in}
    \caption{\textbf{Test on simulation dataset:} The experiment with each configuration is repeated 10 times. Ours outperforms Ours\_woKF both on robustness and accuracy.} 
    \label{fig:vio_test}
    \vspace{-0.2in}
\end{figure}

\textbf{Robustness evaluation:} We use a configuration of a 10-keyframe sliding window and a 10-degree initial error to evaluate the robustness of the methods. The results are shown in Fig. \ref{fig:robust_test} and Tab. \ref{tab:robust_test}. It can be observed that Drt-VIO+VI-BA is more prone to producing poor estimations when the IMU rotates rapidly. By considering the extrinsic orientation, the good estimation ratio of our method exceeds that of Drt-VI-BA by more than 6\%. Additionally, the results demonstrate the effectiveness of the failure detection module, with the proposed detection method keeping the non-detected bad rate below 1\%. We also apply the failure detection module to Drt+IRLS+VI-BA for comparison. However, due to the omission of $\bold{R}_{CI}$, the computed feature pairs' weights are incorrect, resulting in few pairs passing the Chi-square test.

\textbf{Efficiency evaluation:} The average runtime of each module for 10KFs and 10 degrees initial error setting is shown in Tab \ref{tab:efficiency_test}. 
The experiments were conducted on a computer with a 3.7 GHz Intel Core i9 CPU and 32 GB of memory. It can be observed that the total runtime of Ours+VI-BA is slightly slower than Drt+VI-BA, but over 1.4 times faster than VINS+VI-BA. Additionally, due to the more accurate initial values provided for VI-BA, Ours+VI-BA is 5ms faster than others in the VI-BA module. Since the proposed method offers precise rotation estimation, if we ignore the influence of extrinsic translation and accelerometer bias, initialization can be completed in less than 15ms, which is 4.3 times faster than VINS+VI-BA and 2.7 times faster than Drt+VI-BA.

\textbf{Ablation Experiment:} We use the estimated error of $\bold{R}_{C_iC_j}$ in degrees as the comprehensive evaluation indicator for the estimation of $\bold{R}_{CI}$ and $\bold{b}_{g}^{I}$. The results in Fig. \ref{fig:ablation_test} show that the proposed FP weighting strategy is better than the $\lambda$ weighting strategy in improving accuracy. Additionally, when the carrier rotates rapidly, as in sequences V1\_03 and V2\_03, there will be more outliers, and the advantage of the FP weighting strategy will become more pronounced. This is due to the weighting of finer-grained error terms and culling outliers through the Chi-square test.
Comparing the results of ours and ours only FP weight, the difference between the two is not significant and does not reflect the role of IMU pre-integration covariance in weighting. But this is due to the fact that we select keyframes according to a constant frequency. When the keyframe selection strategy is not at a fixed frequency, our method considering IMU pre-integration covariance is recommended.

\subsection{VIO Experiments}
\label{sec:vio_test}
Finally, we evaluate the effect of the proposed VIO initialization on the entire VIO system by replacing the initialization of the VINS-Mono system by Ours+VI-BA and Drt+VI-BA, namely Ours and Drt-VIO. VINS-Mono(Post) and VINS-Mono(Pre) denote the VIO system using the post-estimating method and the pre-estimating method provided by the original VINS-Mono, respectively. 
Besides, Ours\_woKF denotes our VIO system without using the IESKF module in the initialization stage, where only the previous estimation is used as the initial value for the current estimation. All VIO systems are evaluated on the EuRoC dataset and simulation dataset. The RMSE of trajectory serves as the evaluating indicator.

\textbf{Test on the EuRoC dataset:} We sample 20-second-long data segments every 5 seconds on each sequence and test initialization methods on them. The results are shown in Fig. \ref{fig:vio_euroc_test}. It can be seen that our method significantly outperforms VINS-Mono and Drt-VIO, especially on rapidly rotated sequences (V1\_02, V1\_03, V2\_02, and V1\_03).  Since there are few instances of pure or approximately pure rotational motions in the EuRoC dataset, Ours has no notable advantage on trajectory accuracy compared to Ours\_woKF. We will demonstrate this advantage on the simulation dataset.

\textbf{Test on the simulation dataset:} We construct a simulated scene containing pure rotation motion, as shown in Fig. \ref{fig:AR_SIM_TEST}, which contains 500 random points and 0.5-pixel Gaussian noise is introduced to perturb each point observation in images. IMU noise and camera intrinsic are set as the EuRoC datasets \cite{burri2016euroc}. The initial extrinsic orientation error is set to 10 degrees. The total motion time is set to 50 seconds, and we test the initialization performance under various durations of pure rotational motion. The results are shown in Fig. \ref{fig:vio_test}. It can be observed that as the duration of pure rotational motion increases, the advantage of our method compared to Ours\_woKF becomes increasingly evident. By jointly considering the extrinsic orientation and gyroscope bias, our method results in a significant improvement in the trajectory estimation of the entire VIO system.

\section{Conclusions}
This paper proposes an extrinsic orientation and gyroscope bias estimation for visual-inertial initialization. A novel weighting strategy is introduced to enhance the accuracy and robustness of the estimation, and a MAP estimation is applied to refine the results before sufficient translation is achieved. Extensive experiments demonstrate that our method provides more accurate and robust initialization compared to classical and state-of-the-art methods while maintaining competitive efficiency.
Our method overlooks the effect of IMU intrinsic parameters influencing the performance of AR applications on low-end mobile devices, which will be our future work.

\addtolength{\textheight}{-11cm}   

\bibliographystyle{ieeetr}
\bibliography{IEEEabrv, DOGE}

\end{document}